\documentclass[10pt,twocolumn,letterpaper]{article}
\pdfoutput=1 
\usepackage{wacv}
\usepackage{times}
\usepackage{epsfig}
\usepackage{graphicx}
\usepackage{amsmath}
\usepackage{amssymb}
\usepackage{subfigure}
\usepackage{graphics}
\usepackage[utf8]{inputenc} %
\usepackage[T1]{fontenc}    %
\usepackage{url}            %
\usepackage{booktabs}       %
\usepackage{amsfonts}       %
\usepackage{nicefrac}       %
\usepackage{microtype}      %
\usepackage{lipsum}
\usepackage{float}
\usepackage{bm}
\usepackage{xcolor}
\usepackage[ruled,linesnumbered]{algorithm2e}
\usepackage[pagebackref=true,breaklinks=true,colorlinks,bookmarks=false,pdfproducer={}]{hyperref}

\wacvfinalcopy %

\def\wacvPaperID{227} %

\DeclareMathAlphabet{\mathitbf}{\encodingdefault}{\itdefault}{bx}{n}
\ifwacvfinal\fi
\begin{document}

\title{Triangle-Net: Towards Robustness in Point Cloud Learning}

\author{Chenxi Xiao \\
Purdue University\\
{\tt\small xiao237@purdue.edu}
\and
Juan Wachs \\
Purdue University\\
{\tt\small jpwachs@purdue.edu}
}

\maketitle
\ifwacvfinal\thispagestyle{empty}\fi

\begin{abstract}
Three dimensional (3D) object recognition is becoming a key desired capability for many computer vision systems such as autonomous vehicles, service robots and surveillance drones to operate more effectively in unstructured environments. These real-time systems require effective classification methods that are robust to various sampling resolutions, noisy measurements, and unconstrained pose configurations. 
Previous research has shown that points' sparsity, rotation and positional inherent variance can lead to a significant drop in the performance of point cloud based classification techniques. 
{However, neither of them is sufficiently robust to multifactorial variance and significant sparsity. 
In this regard, we propose a novel approach for 3D classification that can simultaneously achieve invariance towards rotation, positional shift, scaling, and is robust to point sparsity. } To this end, we introduce a new feature that utilizes graph structure of point clouds, which can be learned end-to-end with our proposed neural network to acquire a robust latent representation of the 3D object. We show that such latent representations can significantly improve the performance of object classification and retrieval tasks when points are sparse. Further, we show that our approach outperforms PointNet and 3DmFV by 35.0\% and 28.1\% respectively in ModelNet 40 classification tasks using sparse point clouds of only 16 points under arbitrary SO(3) rotation.

\end{abstract}

\section{Introduction}

\begin{figure}[ht]
    \vspace{-2mm}
    \centering
    \includegraphics[width=13mm, height=13mm,trim=120mm 150mm 120mm 100mm, clip=true]{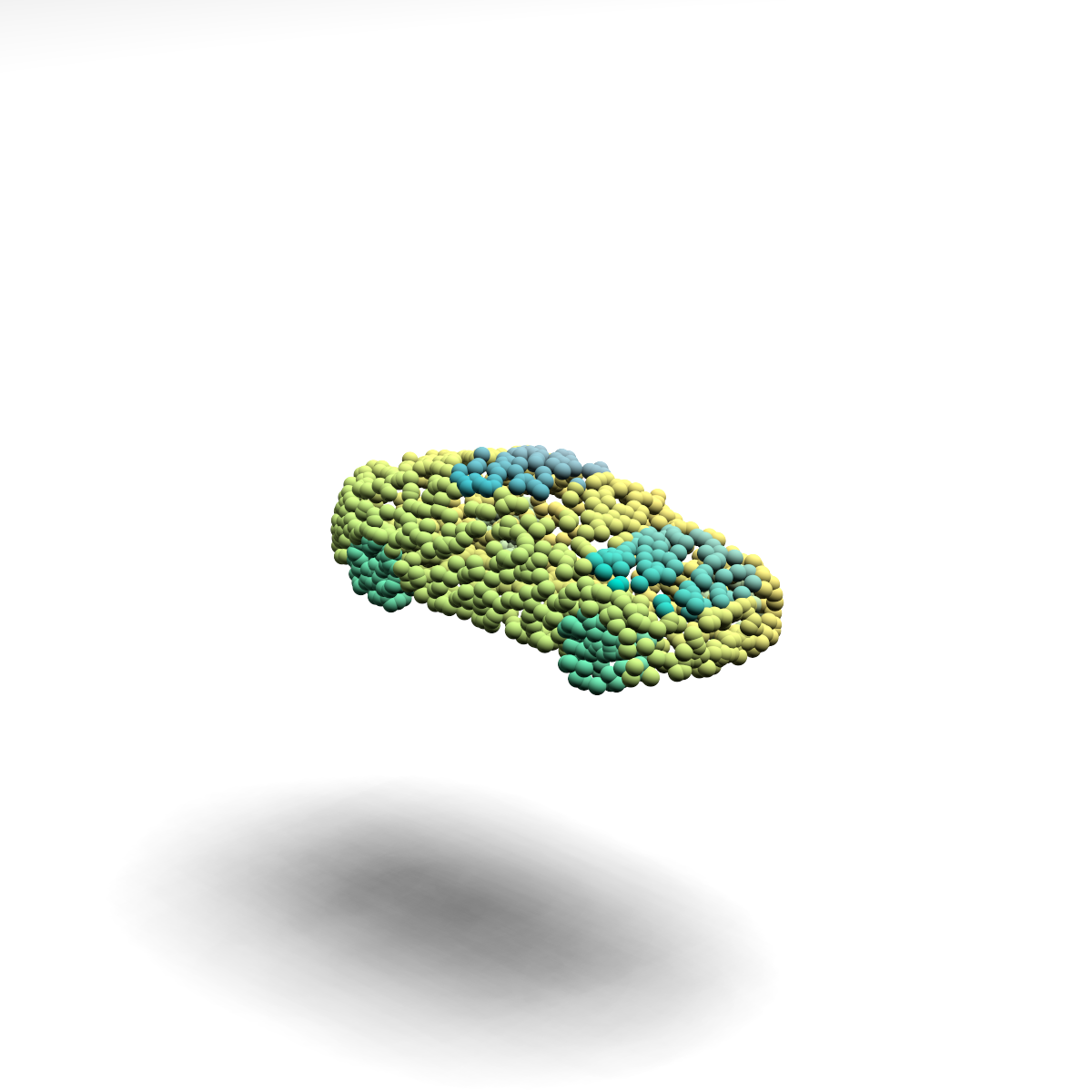} 
    \includegraphics[width=13mm, height=13mm,trim=120mm 150mm 120mm 100mm, clip=true]{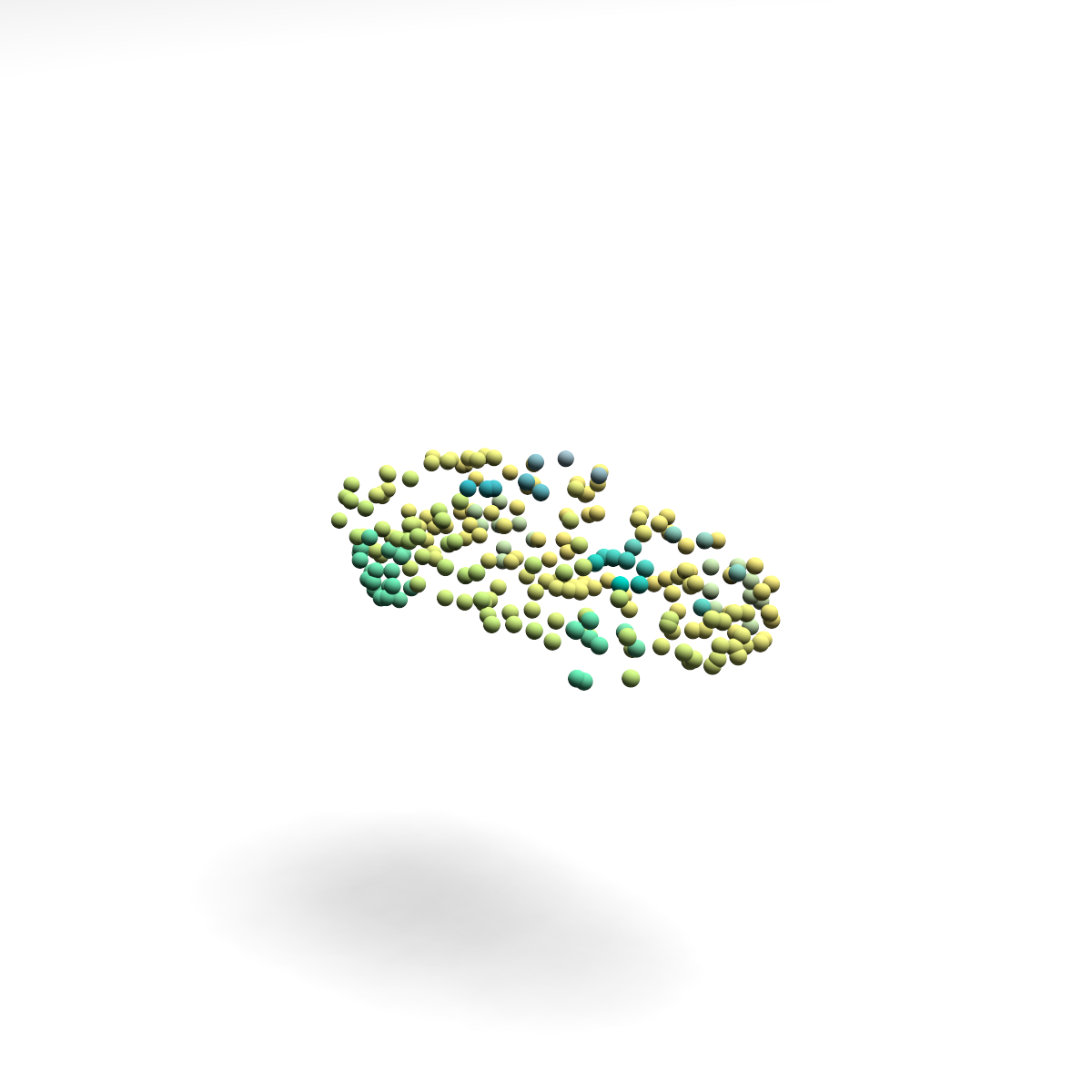} 
    \includegraphics[width=13mm, height=13mm,trim=120mm 150mm 120mm 100mm, clip=true]{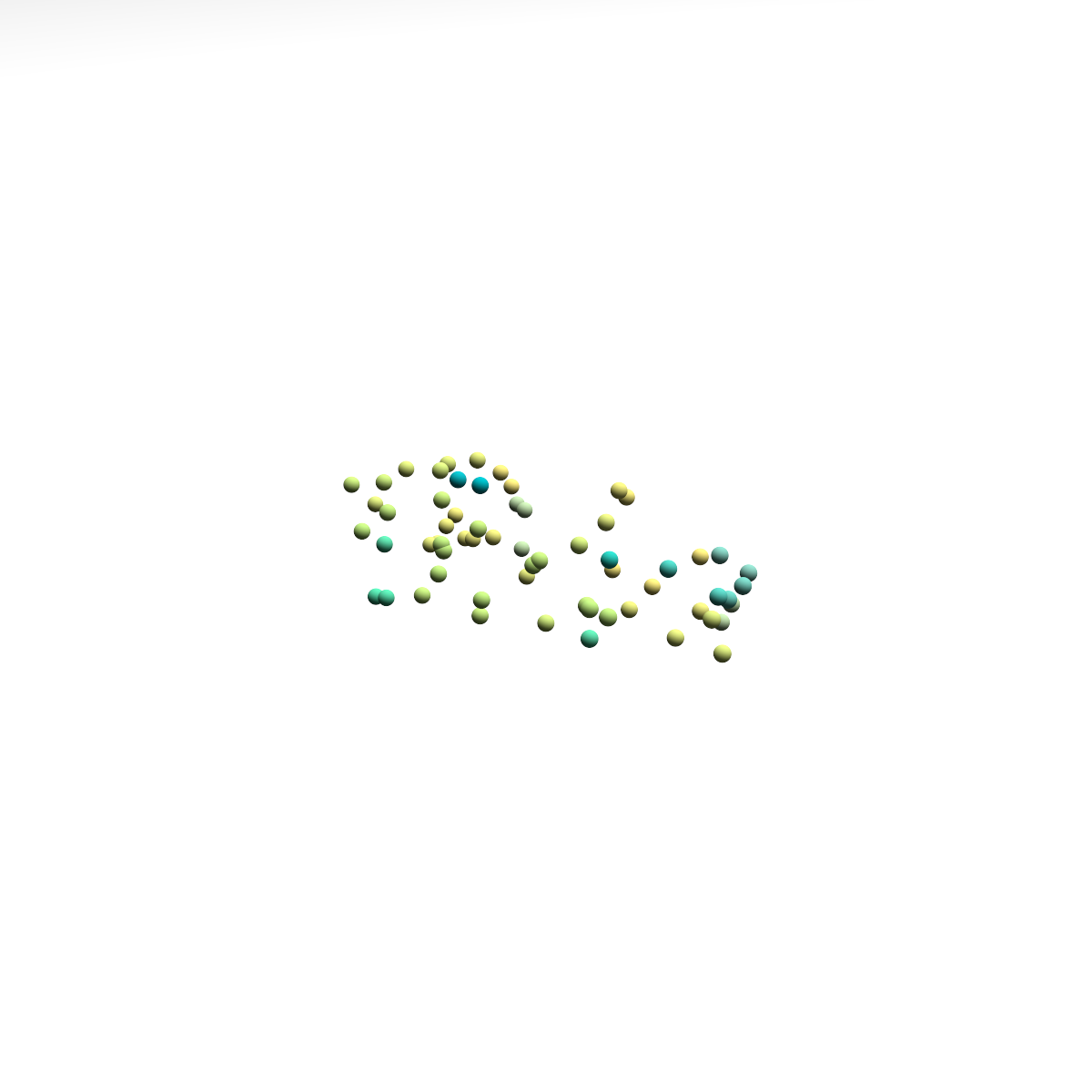}
    \includegraphics[width=13mm, height=13mm,trim=120mm 150mm 120mm 100mm, clip=true]{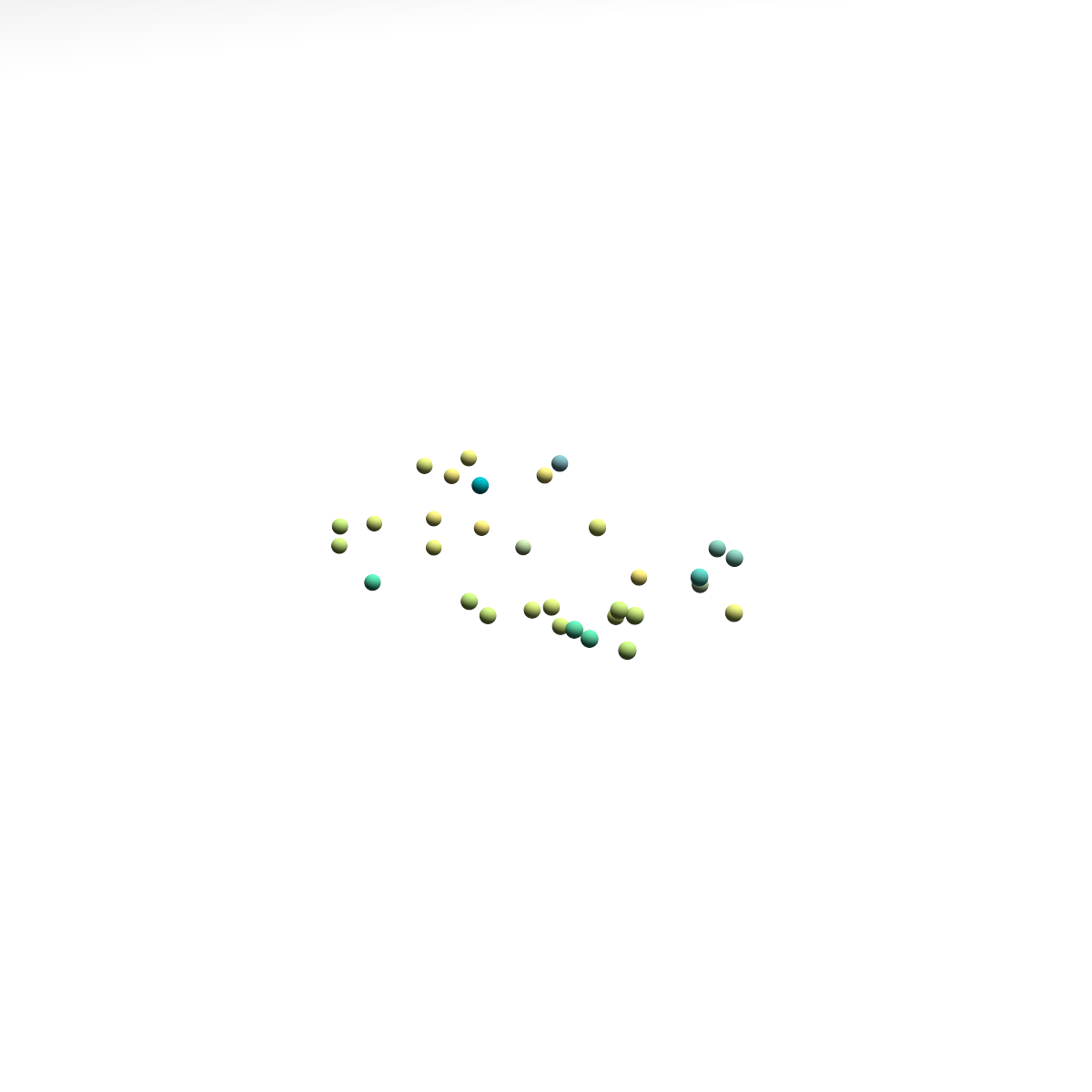}
    \includegraphics[width=13mm, height=13mm,trim=120mm 150mm 120mm 100mm, clip=true]{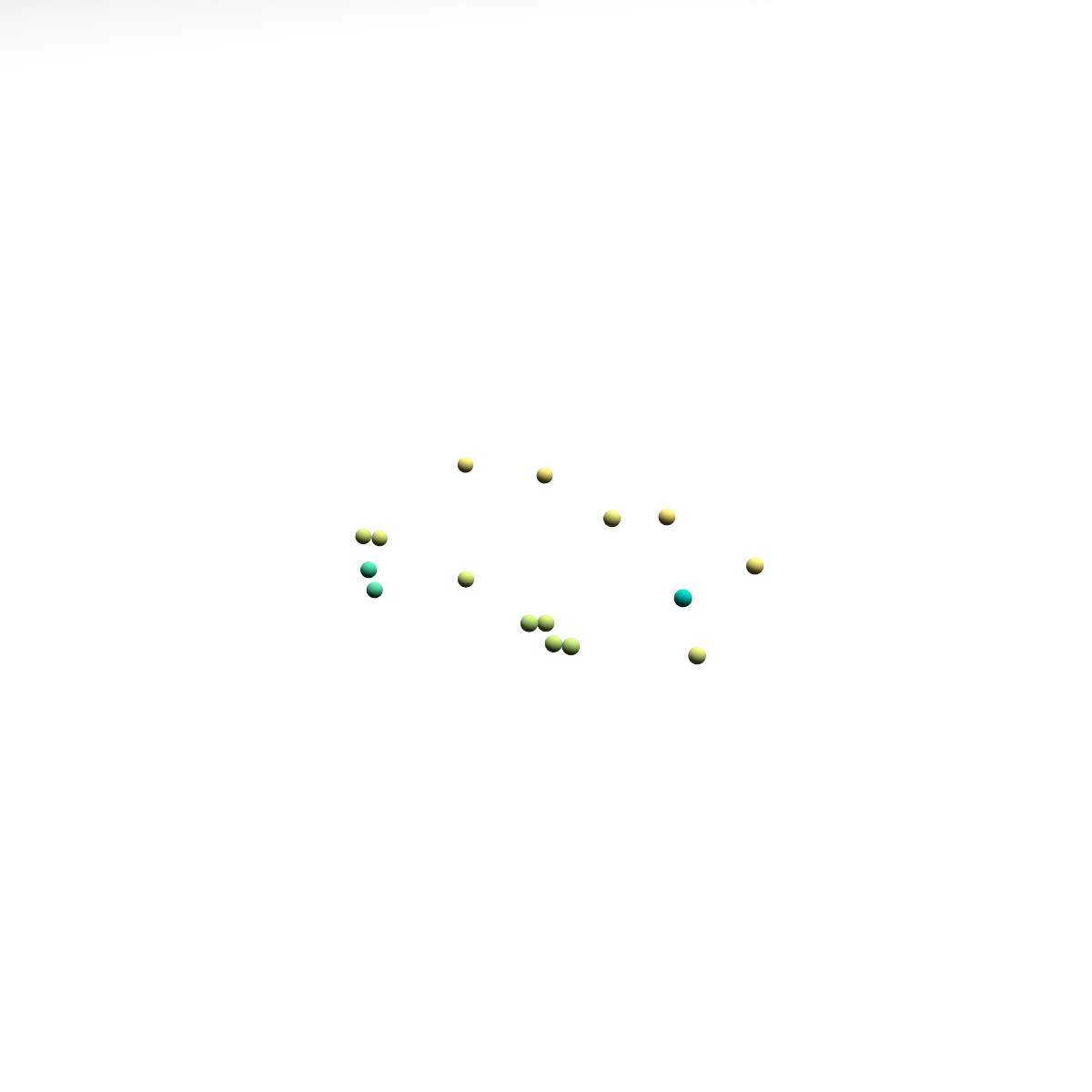}
    \includegraphics[width=13mm, height=13mm,trim=120mm 150mm 120mm 100mm, clip=true]{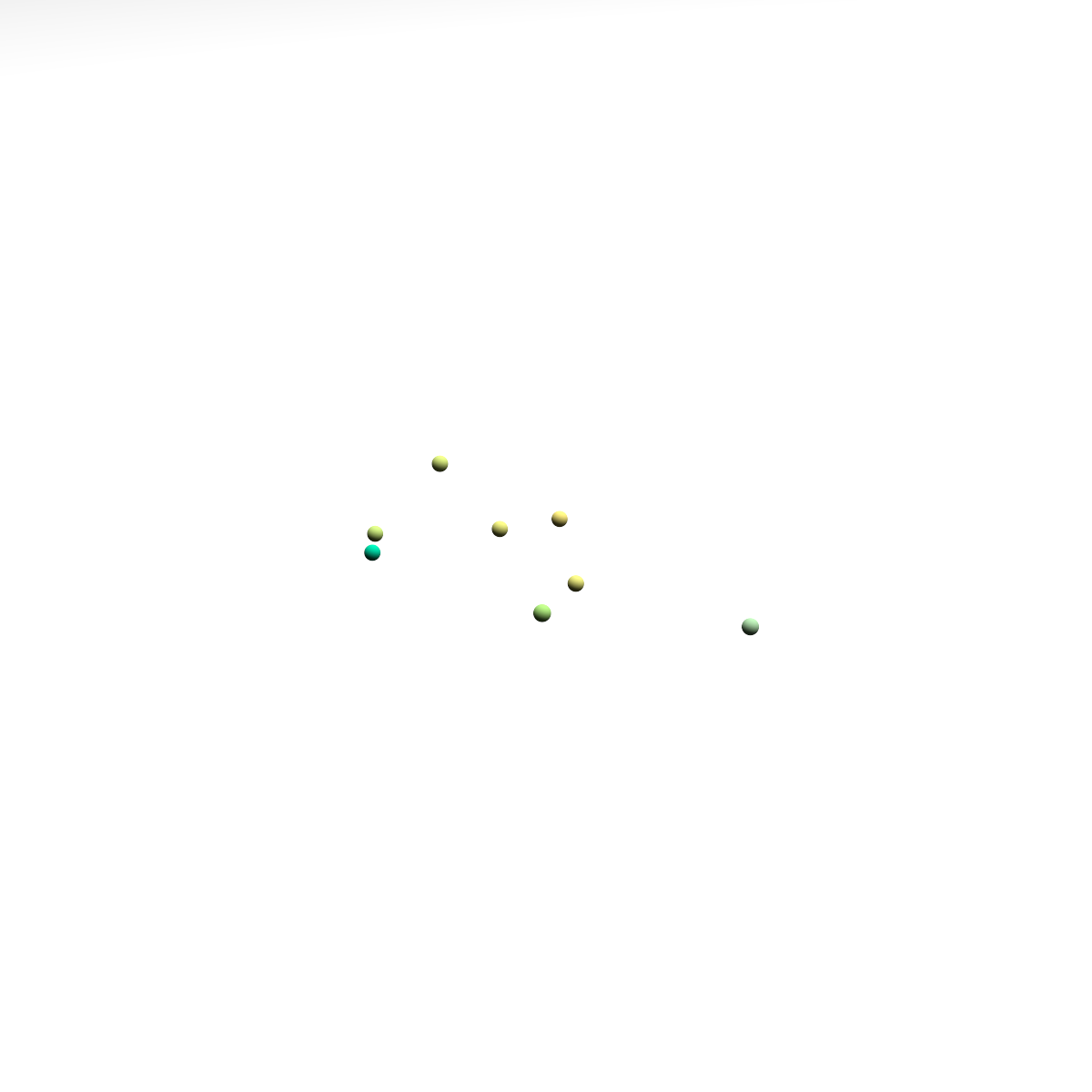}
    
    \includegraphics[width=13mm, height=13mm,trim=120mm 150mm 120mm 160mm, clip=true]{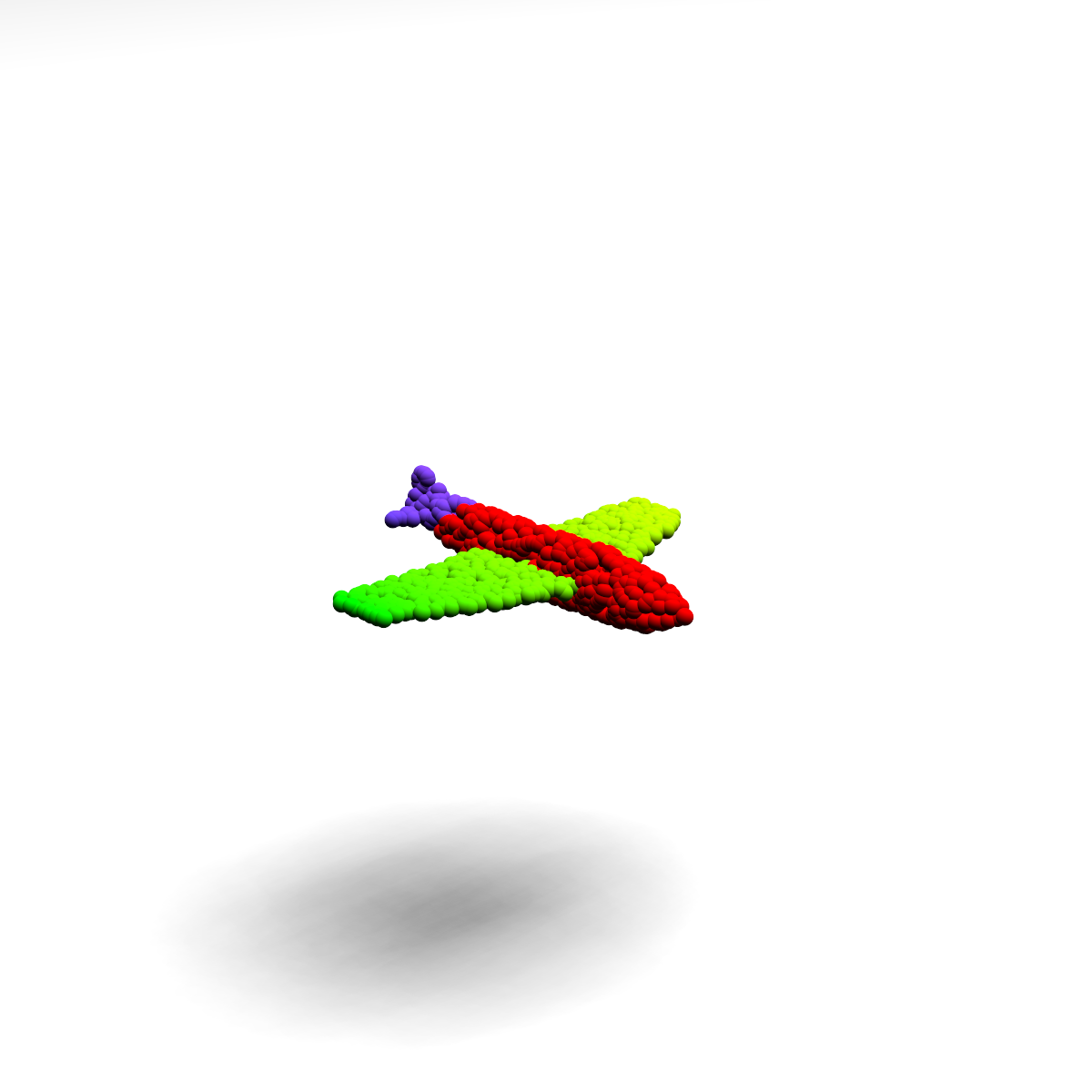} 
    \includegraphics[width=13mm, height=13mm,trim=120mm 150mm 120mm 160mm, clip=true]{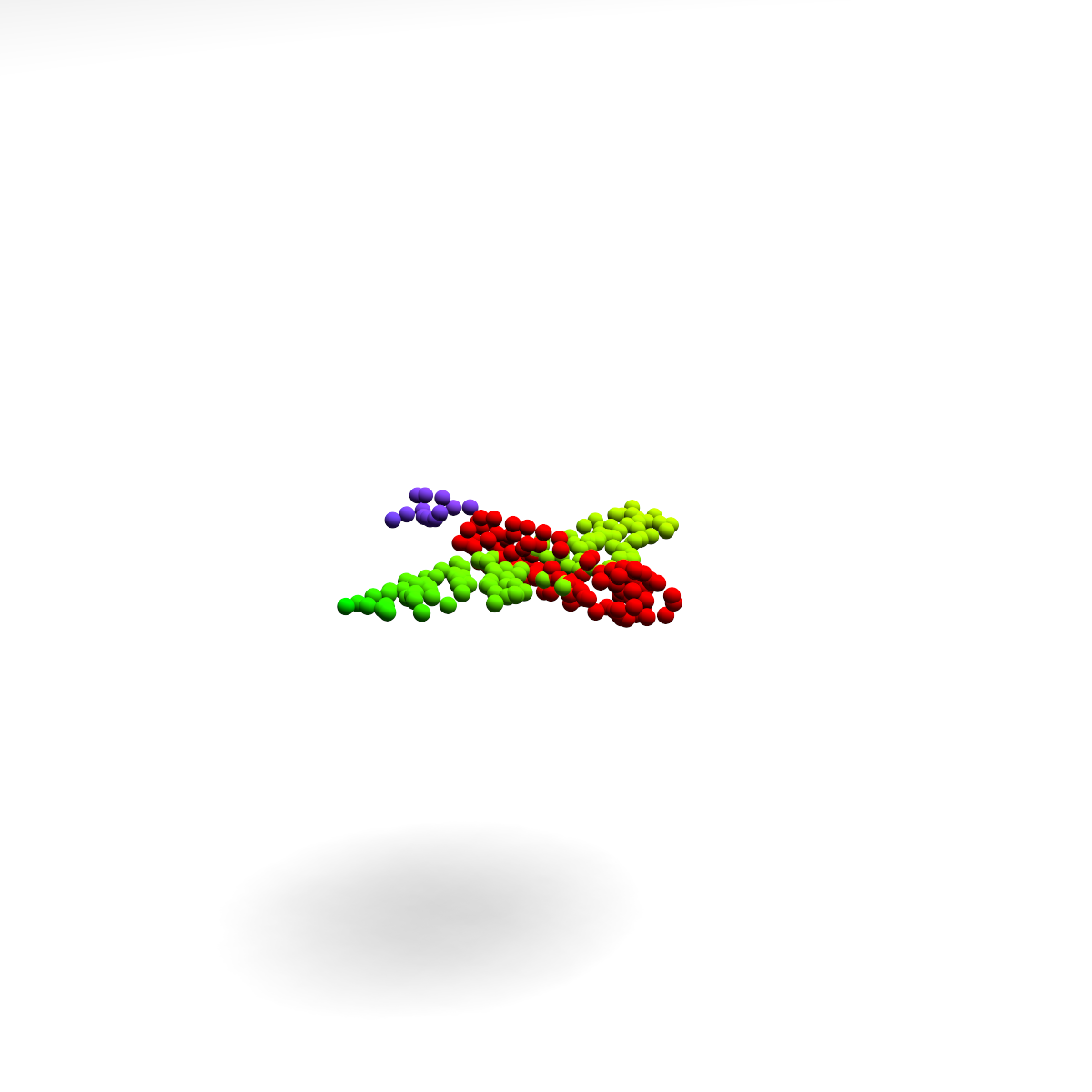} 
    \includegraphics[width=13mm, height=13mm,trim=120mm 150mm 120mm 160mm, clip=true]{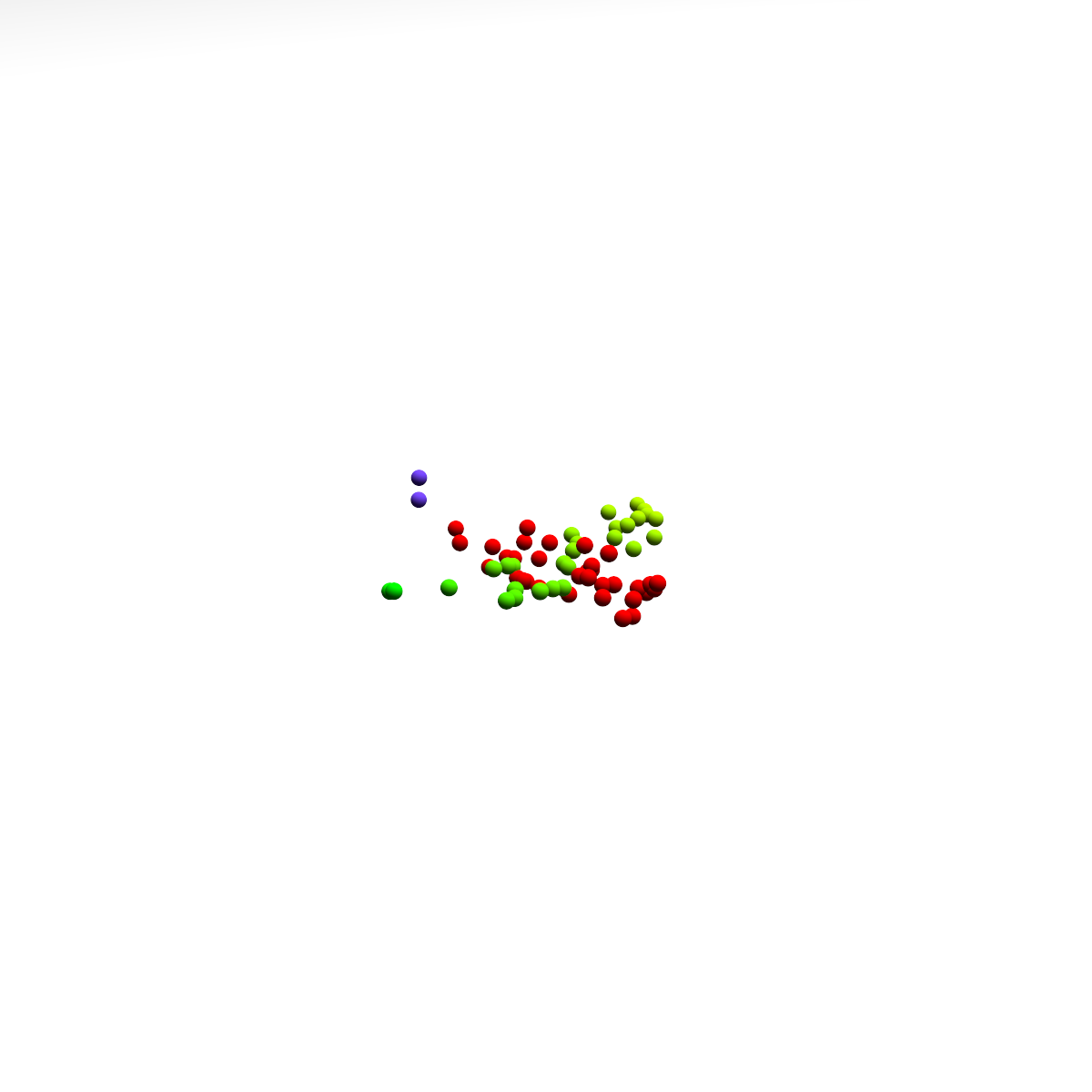}
    \includegraphics[width=13mm, height=13mm,trim=120mm 150mm 120mm 160mm, clip=true]{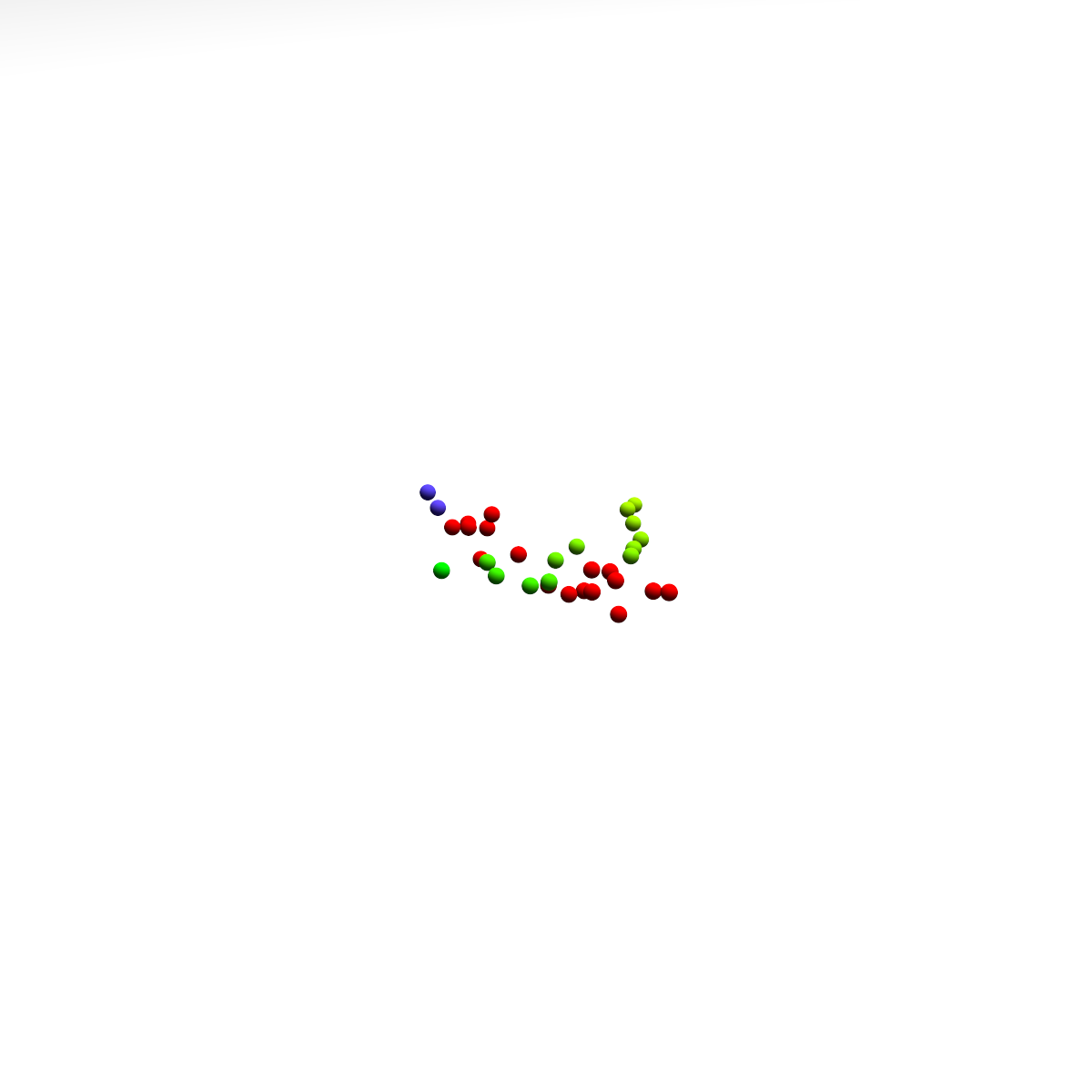}
    \includegraphics[width=13mm, height=13mm,trim=120mm 150mm 120mm 160mm, clip=true]{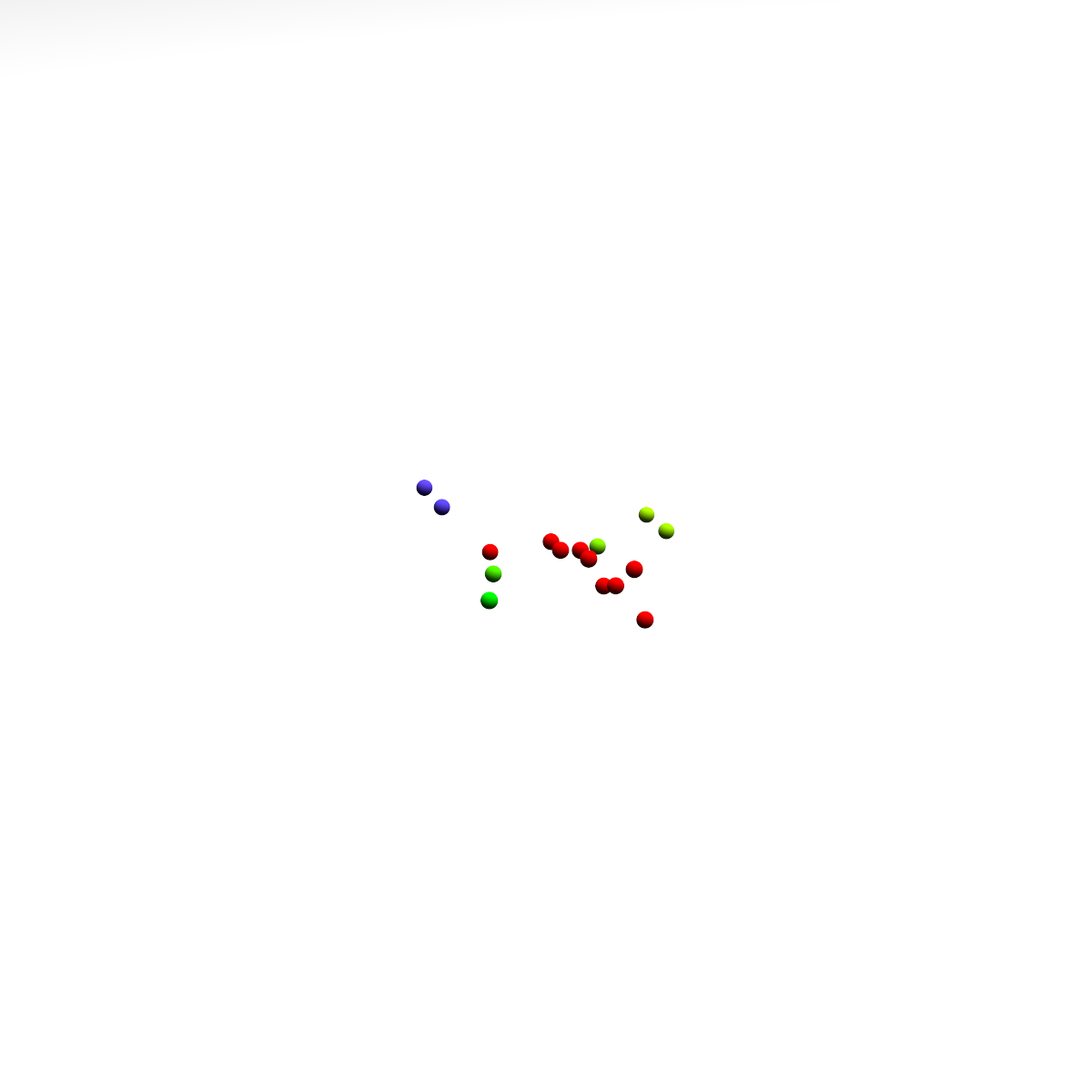}
    \includegraphics[width=13mm, height=13mm,trim=120mm 150mm 120mm 160mm, clip=true]{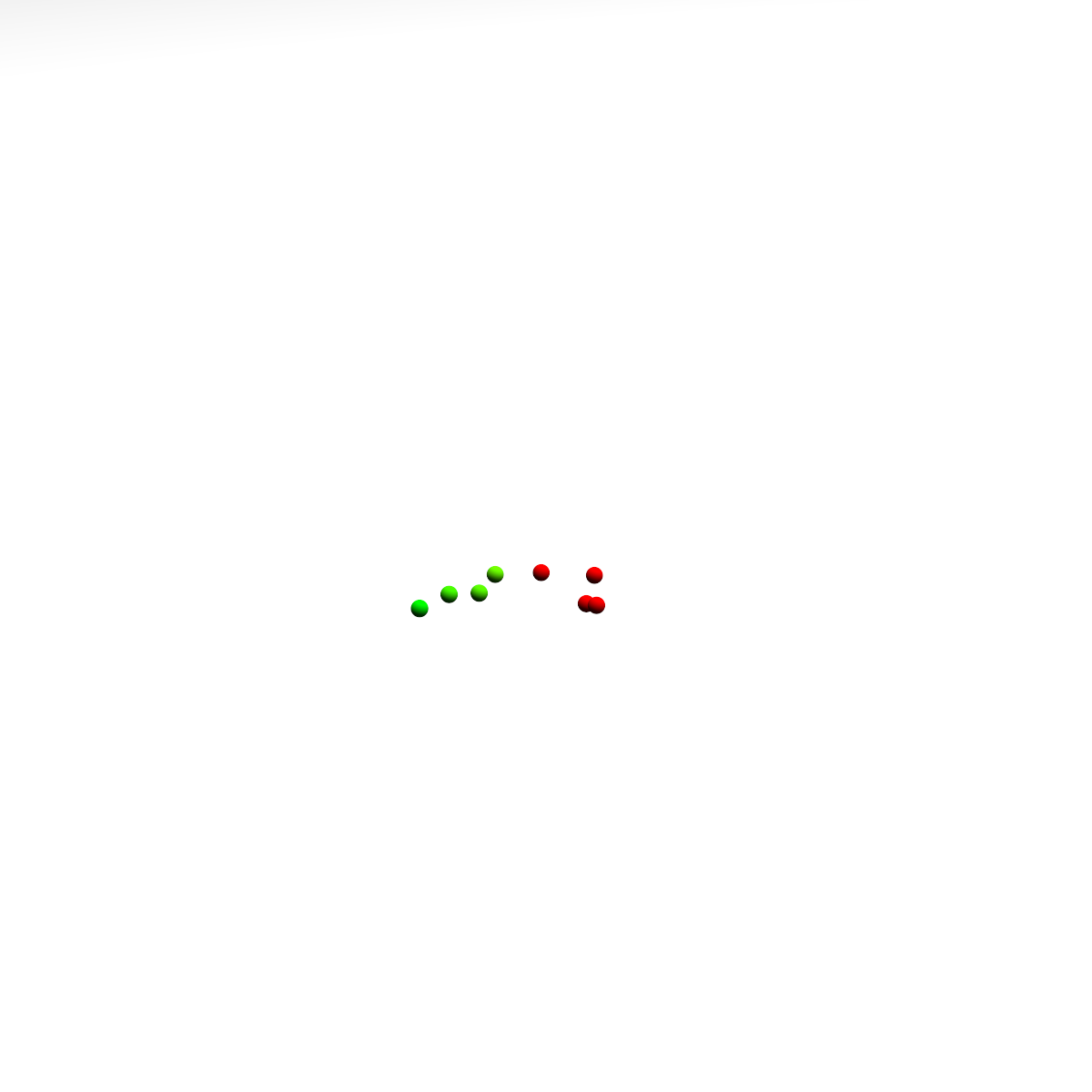}
    
    \includegraphics[width=13mm, height=13mm,trim=120mm 40mm 120mm 200mm, clip=true]{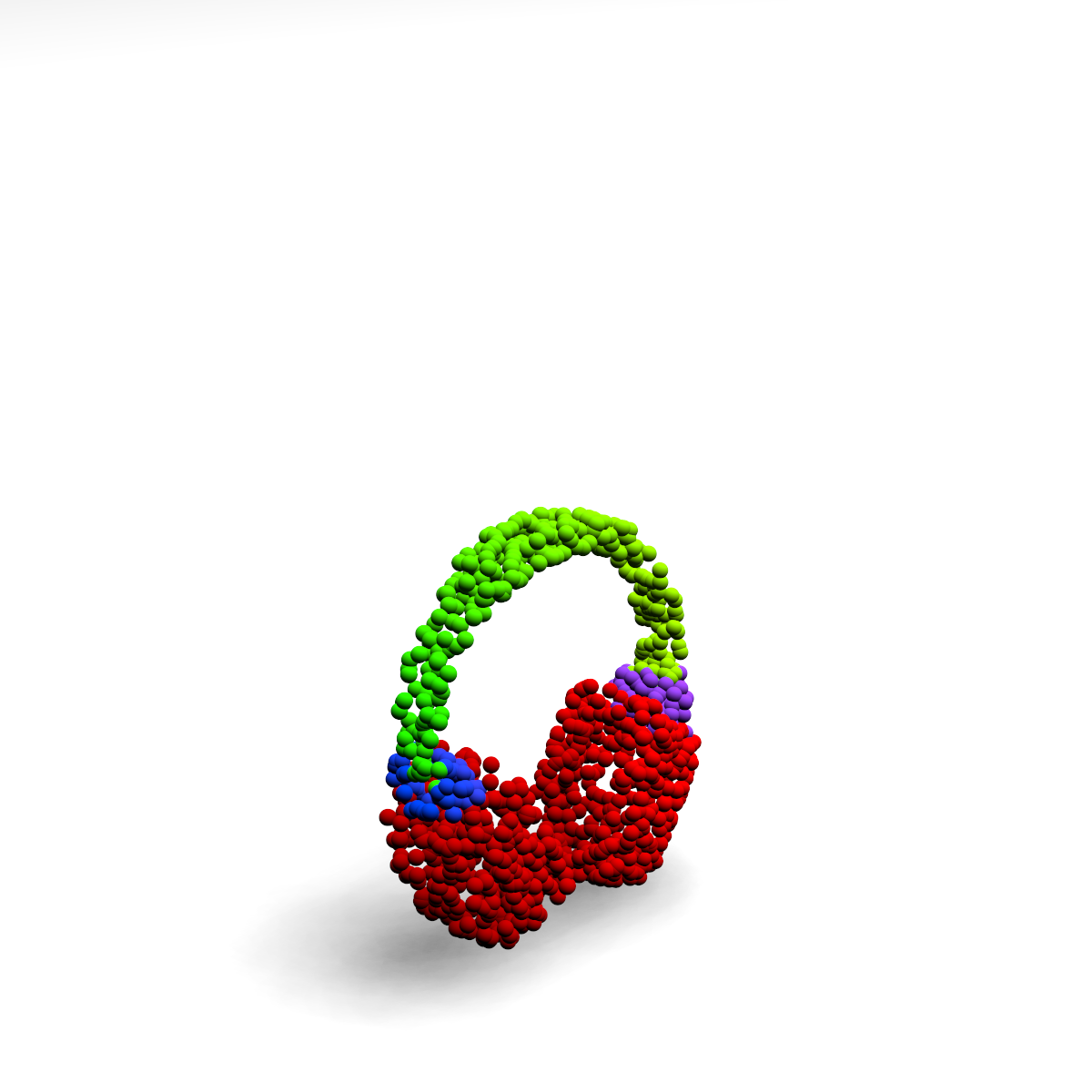}
    \includegraphics[width=13mm, height=13mm,trim=120mm 40mm 120mm 200mm, clip=true]{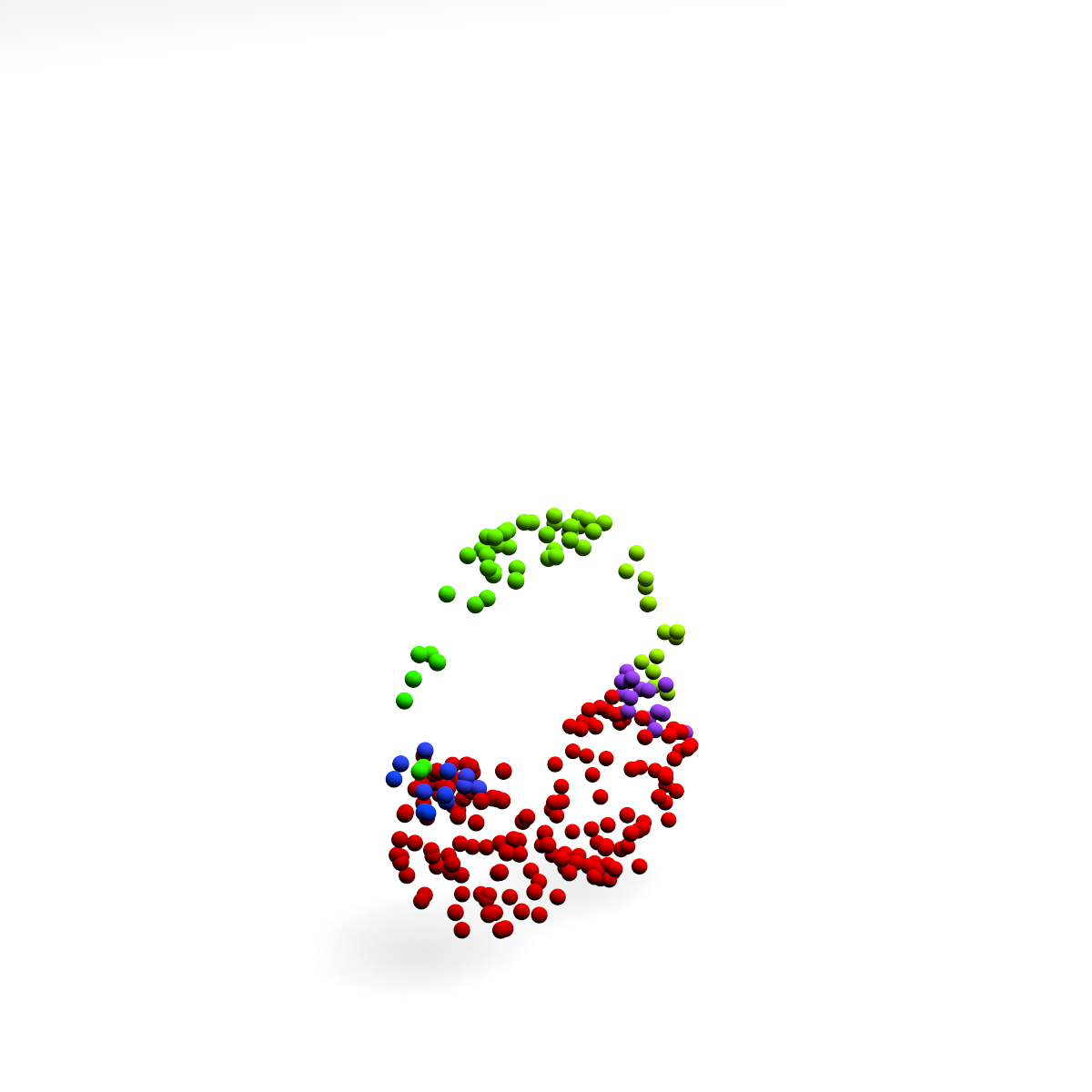}
    \includegraphics[width=13mm, height=13mm,trim=120mm 40mm 120mm 200mm, clip=true]{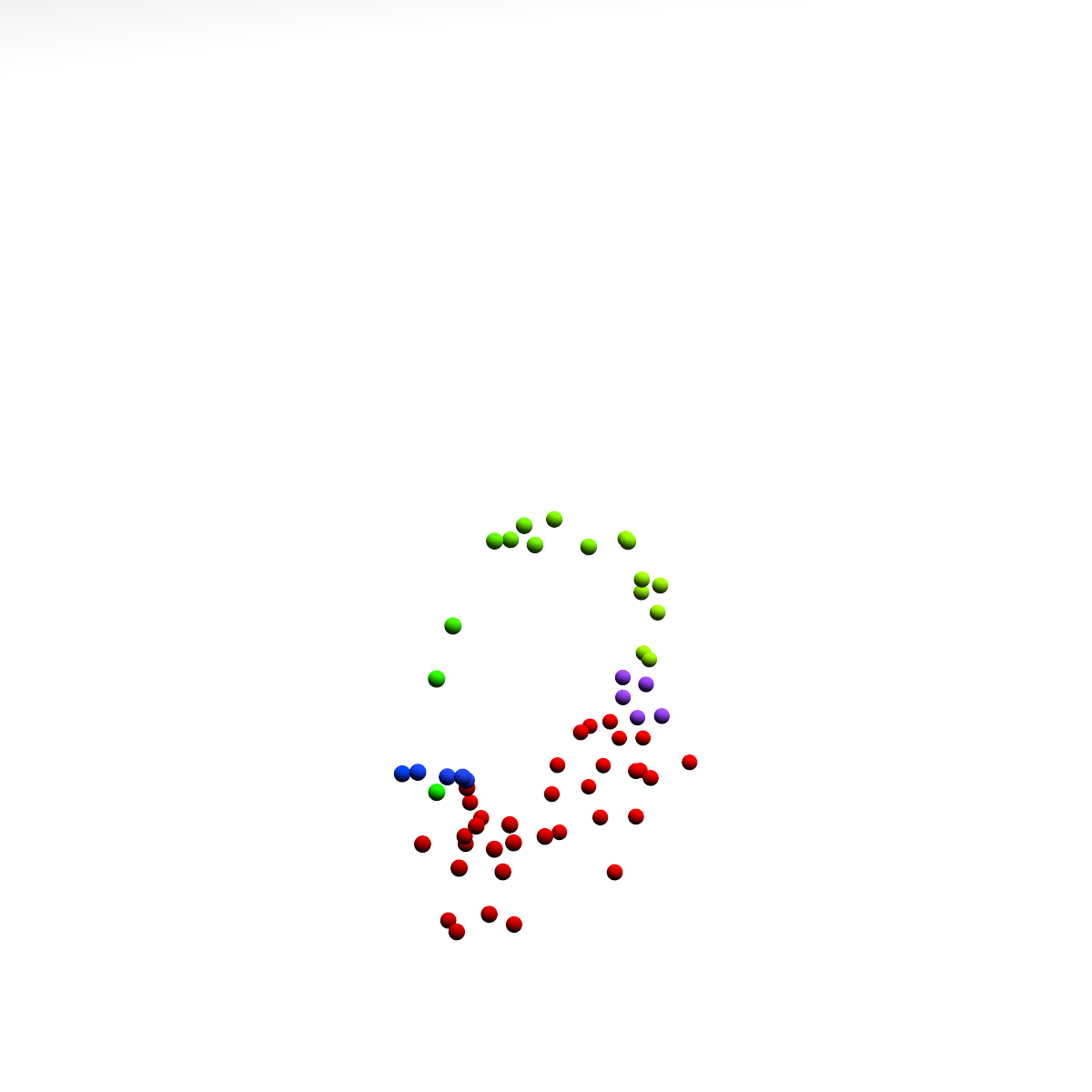}
    \includegraphics[width=13mm, height=13mm,trim=120mm 40mm 120mm 200mm, clip=true]{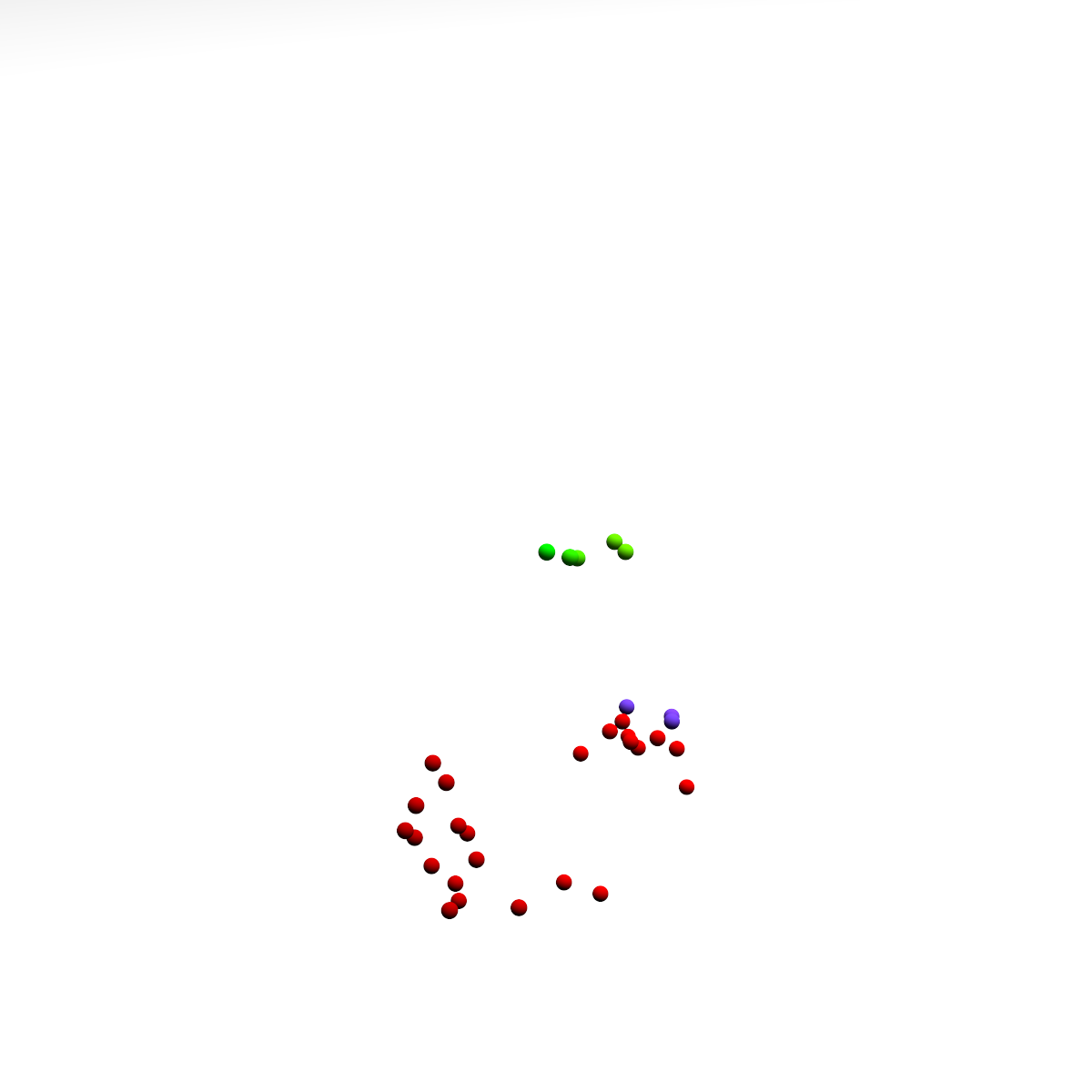}
    \includegraphics[width=13mm, height=13mm,trim=120mm 40mm 120mm 200mm, clip=true]{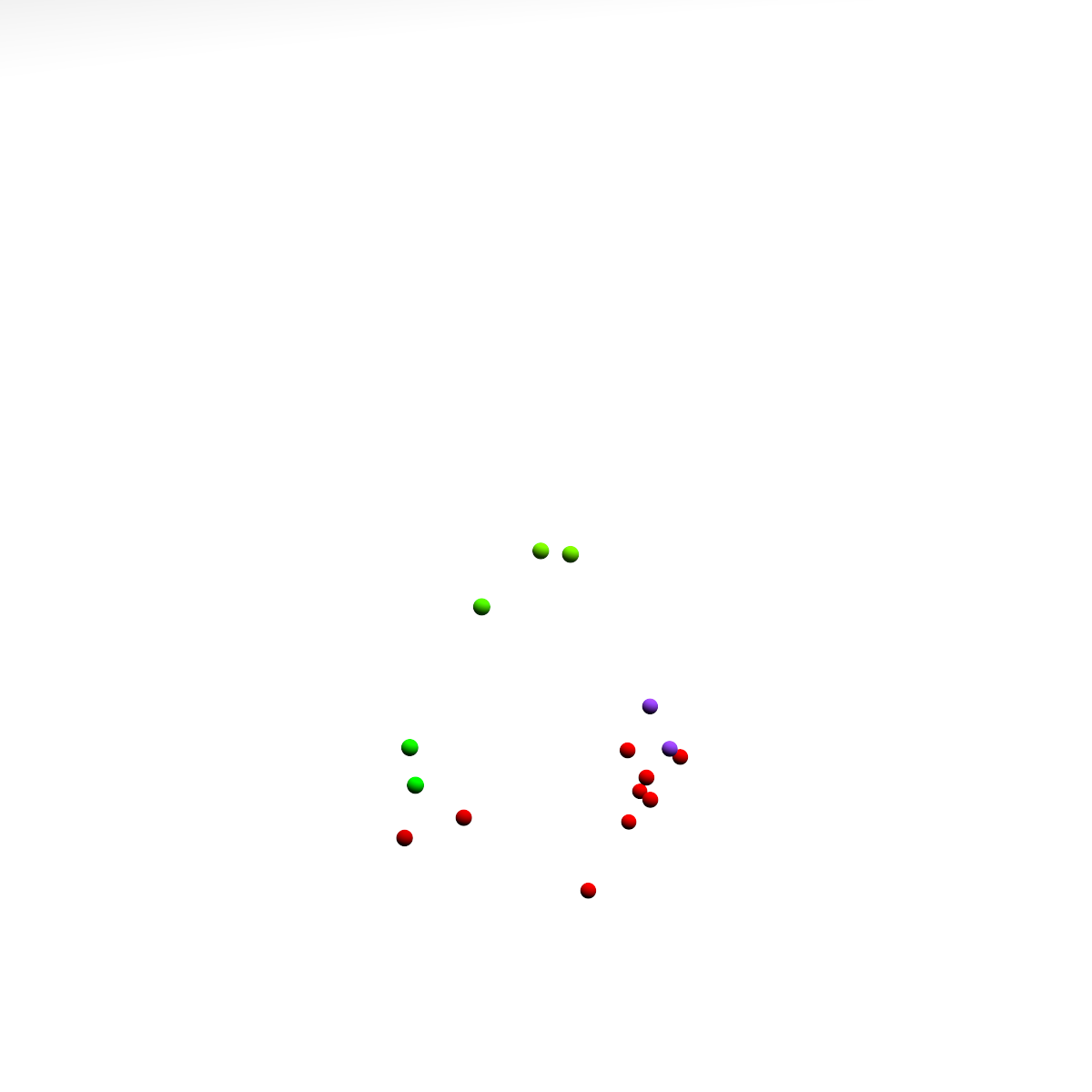}
    \includegraphics[width=13mm, height=13mm,trim=120mm 40mm 120mm 200mm, clip=true]{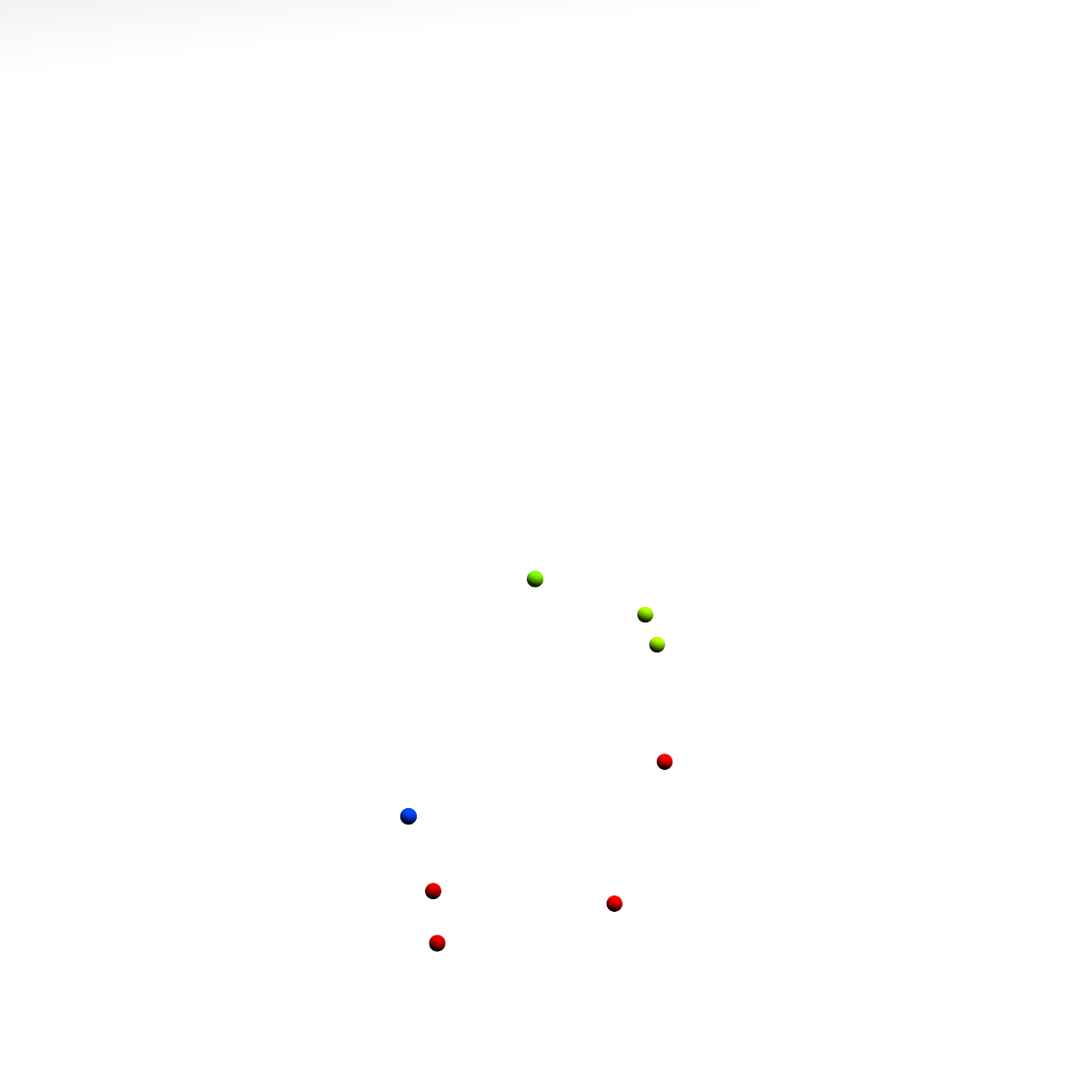}
    
    \includegraphics[width=13mm, height=13mm,trim=120mm 60mm 120mm 150mm, clip=true]{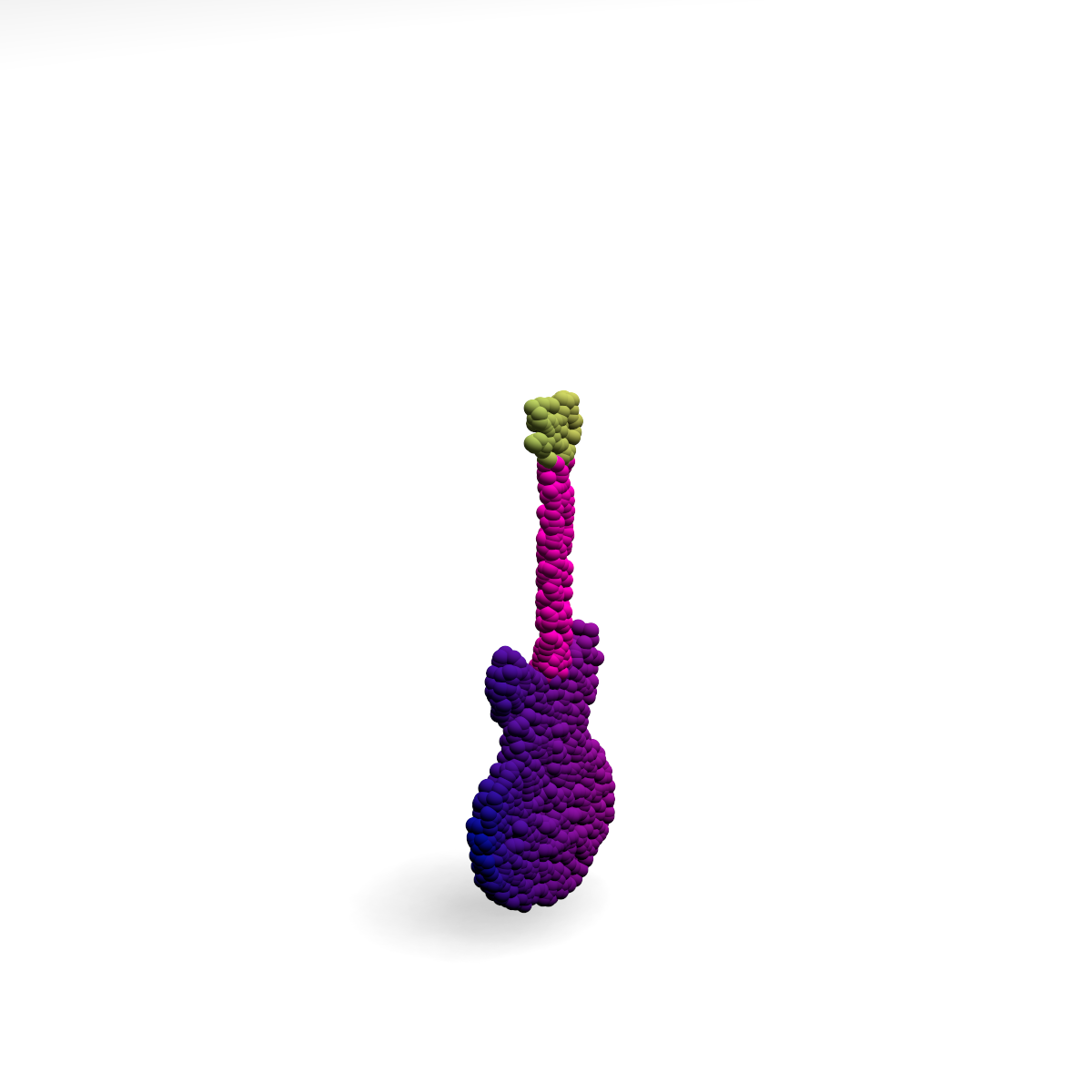}
    \includegraphics[width=13mm, height=13mm,trim=120mm 60mm 120mm 150mm, clip=true]{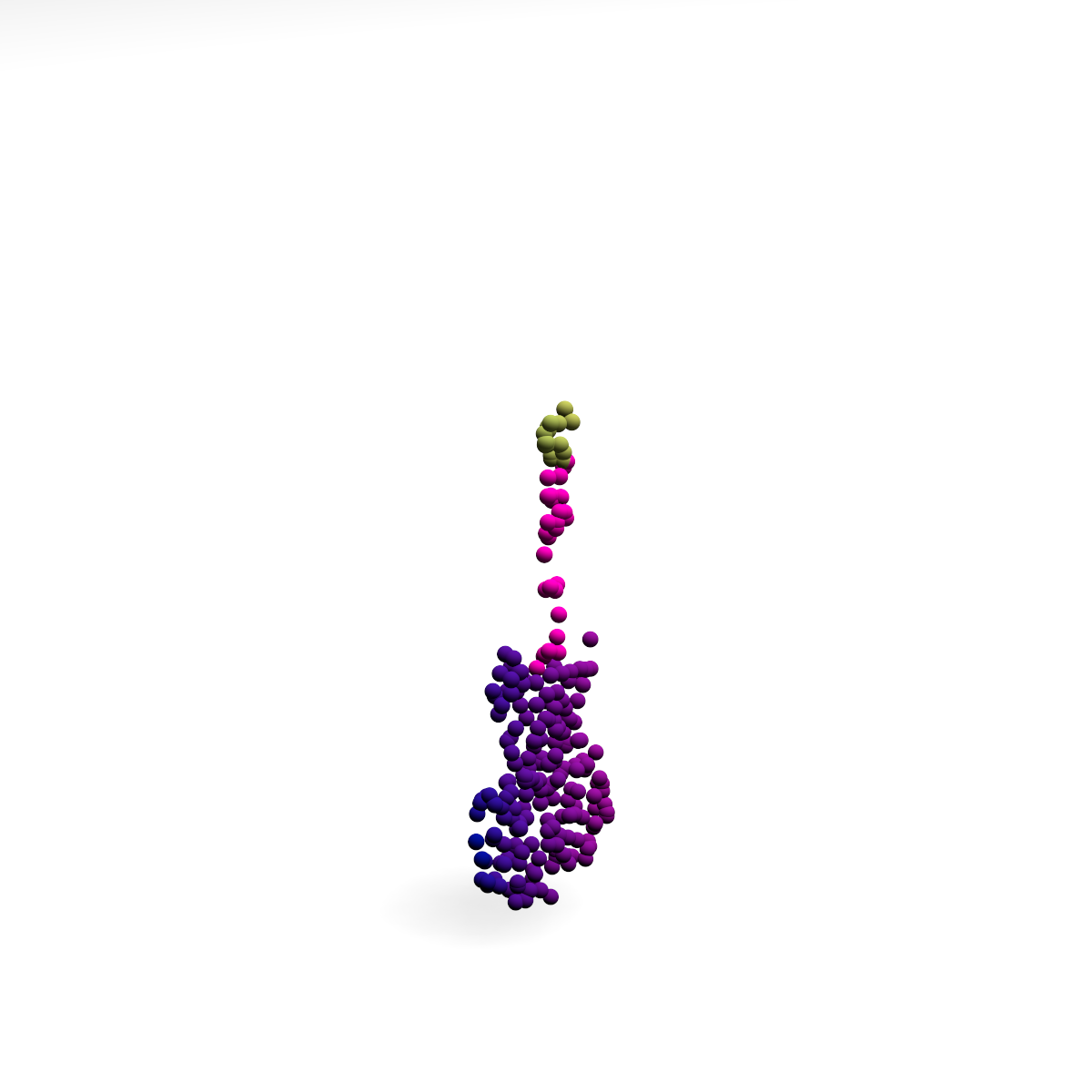}
    \includegraphics[width=13mm, height=13mm,trim=120mm 60mm 120mm 150mm, clip=true]{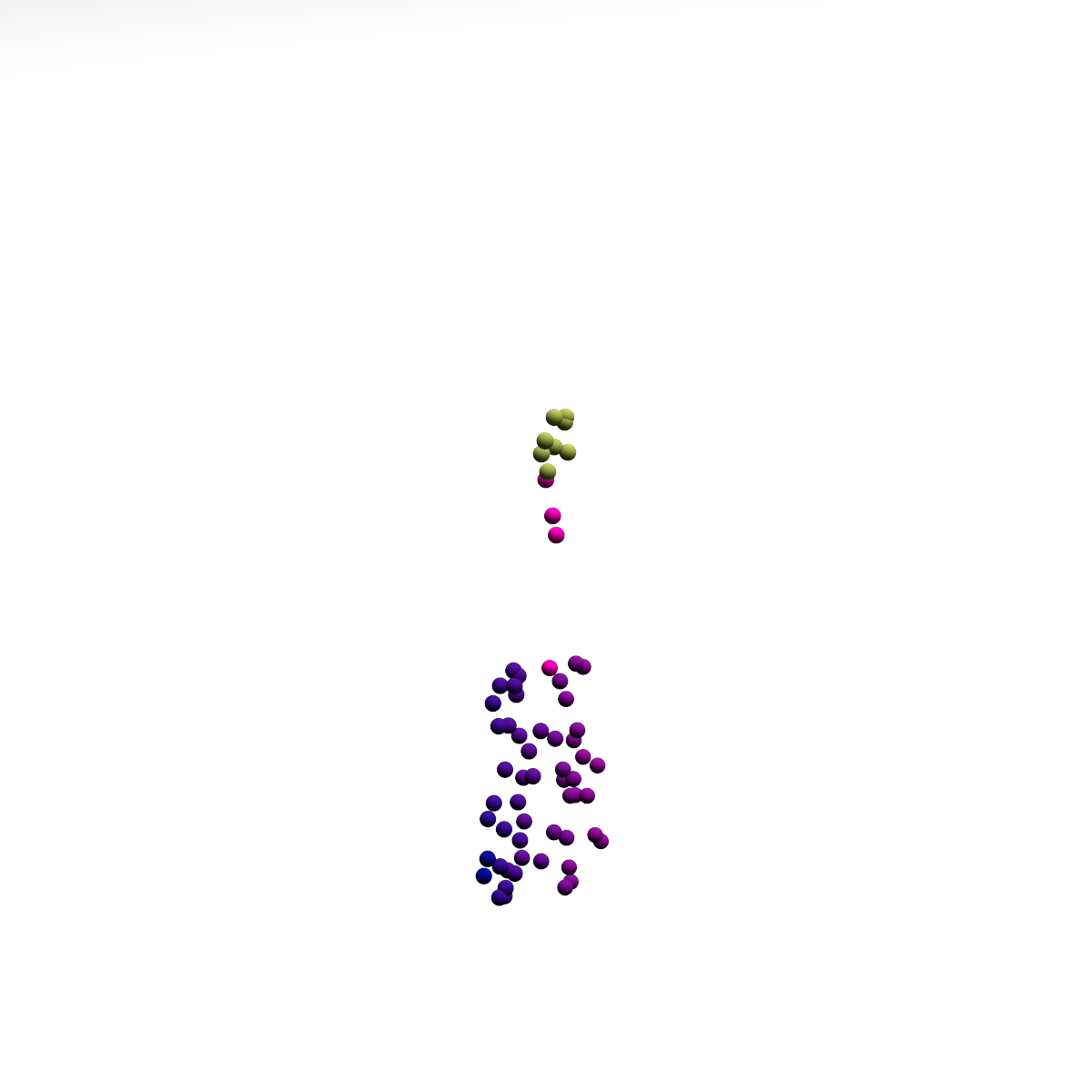}
    \includegraphics[width=13mm, height=13mm,trim=120mm 60mm 120mm 150mm, clip=true]{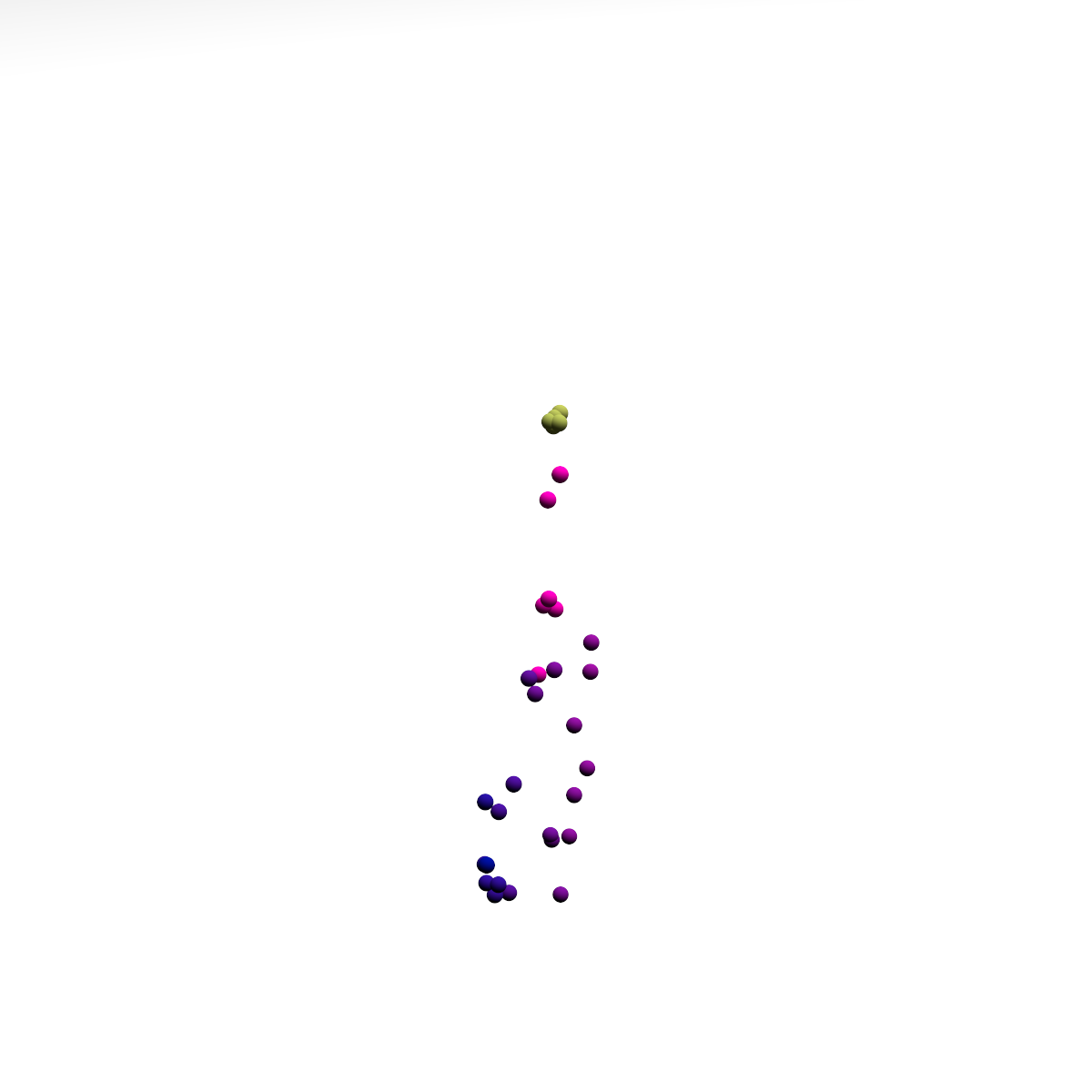}
    \includegraphics[width=13mm, height=13mm,trim=120mm 60mm 120mm 150mm, clip=true]{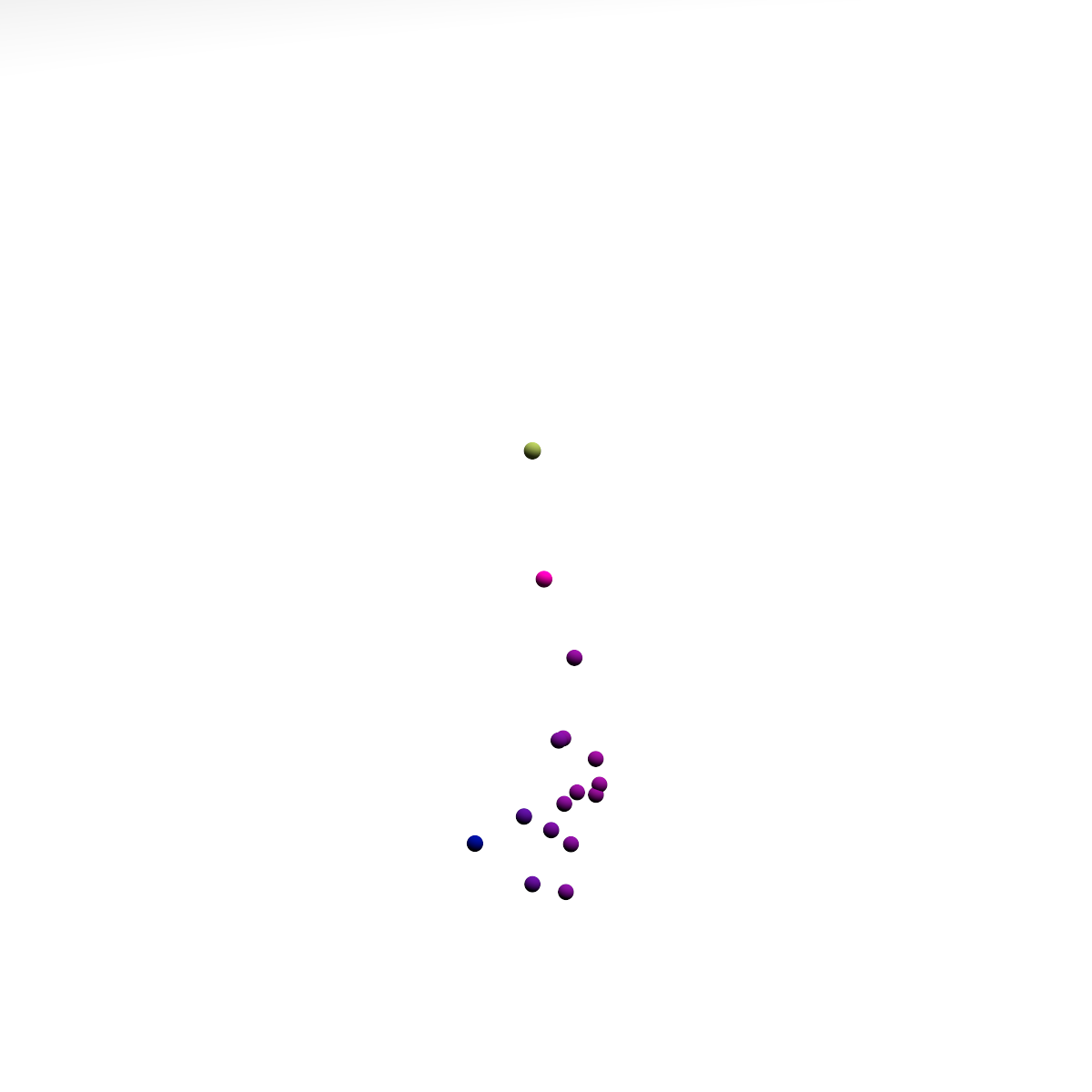}
    \includegraphics[width=13mm, height=13mm,trim=120mm 60mm 120mm 150mm, clip=true]{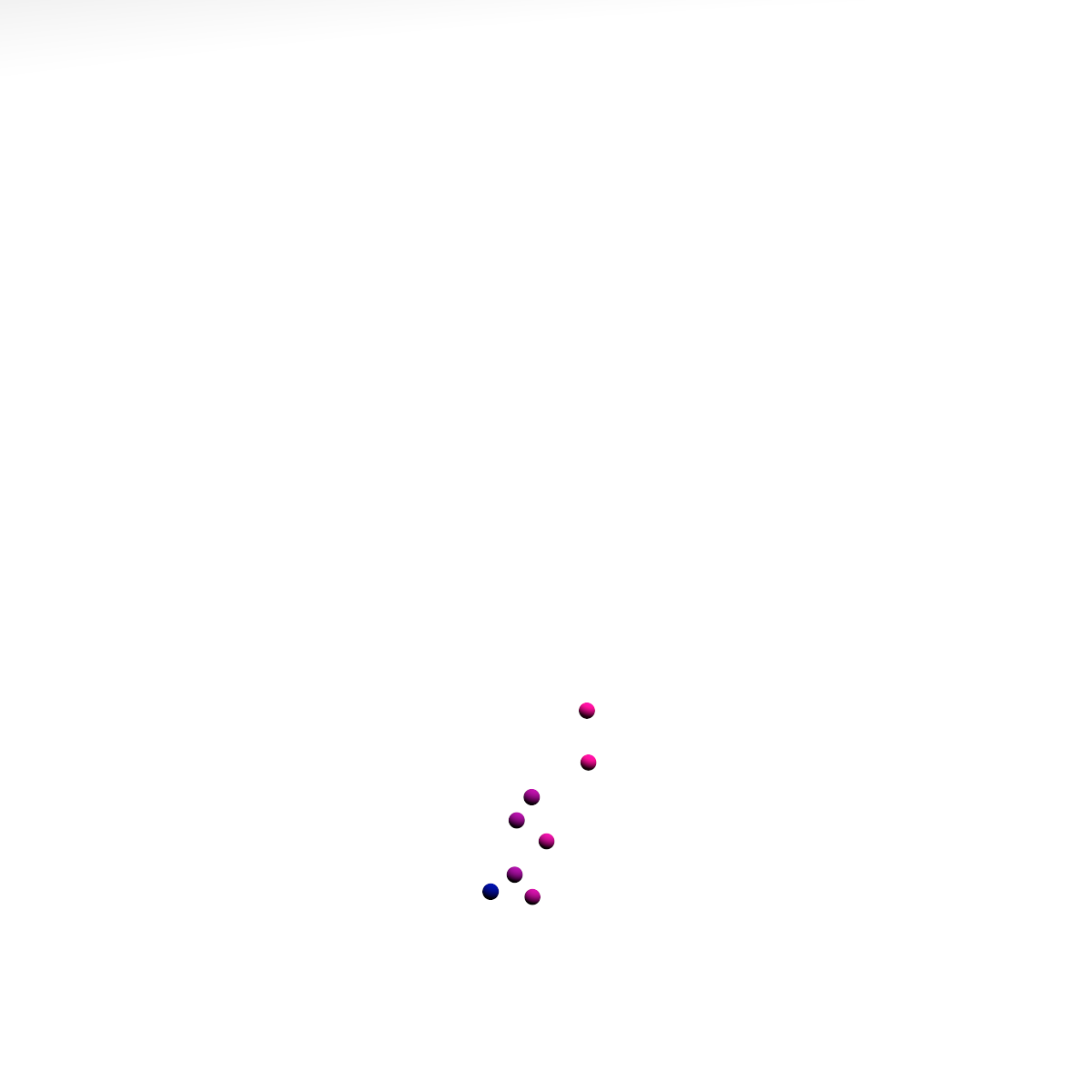}
 
    \caption{Point cloud of car, airplane, earphone and guitar model with size 2048, 256, 64, 32, 16 and 8 (from left to right). The object recognition is challenging even for human when the point cloud size is smaller than 16. Our developed algorithm can recognize objects from 40 categories at 70.35\% using 16 points and 48.19\% using 8 points even under arbitrary SO(3) rotation.}%
    \label{fig:conceptual}
    \vspace{-5mm}
\end{figure}

As commodity cameras and laser based sensors become more affordable, point cloud based object classification is becoming the default approach for 3D sensing. For example, autonomous vehicles rely on point cloud maps sampled by Lidar sensors or depth cameras for effective navigation. One challenge often faced in such applications is that the density of sampling points decreases significantly as the distance from the vehicle embedded sensor to the object increases. This makes it hard to recognize objects that are far from such sensors due to their sparse inherent point structure \cite{bogoslavskyi2017efficient}. As reported in the literature \cite{qi2017pointnet, qi2017pointnet++, ben20183dmfv}, the classification accuracy of these algorithms drop radically as the density of the point cloud decreases, and is further affected when the pose configuration of the object is not known in advance. Similarly, consider the scenario of tactile based object recognition using a robotic hand. The time complexity of sampling is proportional to the number of points sampled along the manipulator's trajectory \cite{zhang2017active, rosales2018gpatlasrrt, elbanhawi2014sampling}. This implies that in addition to performance degradation, there is an additional cost related to the amount of sampling required to make an acceptable prediction. Thus, it is necessary to come up with new 3D machine learning techniques that can classify objects based on “limited” sparse point cloud data and that can operate in real-time, whether for effective navigation (e.g. autonomous driving case) or for user's meaningful perception (e.g. tactile sampling).

{
Unfortunately, object recognition on sparse point clouds with pose uncertainty has not been well addressed. To convey this issue visually, refer to Fig.~\ref{fig:conceptual}. Most of the machine learning techniques are designed for objects with at least 1024 points, in which features are quite distinguishable \cite{PointGCN, qi2017pointnet, qi2017pointnet++, DGCNN, zhao20193d, li2018pointcnn}. Conversely, finding salient features from sparse points is extremely difficult because salient features such as corners, edges and wrinkles are hard to discern with fewer than 128 points; The shape envelop becomes indistinguishable even for humans when the number of points decreases to 16 and below. Further, the ambiguity is aggravated when arbitrary SO(3) rotation is involved. \cite{srivatsan2020sparse} shows how two objects can be nonrigidly aligned after rotation, even they are from two different categories. This calls for a new solution for recognizing sparse points with high discriminability under pose uncertainty.

Although some pioneer point cloud recognition methods have been focusing on the robustness to sparsity \cite{ben20183dmfv, yan2020pointasnl} and pose variance \cite{zhang2019rotation, chen2019clusternet, li2020rotation}, almost all previous works regard the sparsity and pose variance as two independent tasks and fail to consider them as a whole. However, in real applications, different variations are generally combined. For example,  applications such as tactile recognition \cite{zhang2016triangle} or low-resolution outdoor 3D scans (e.g. Sydney Urban Objects, involving point clouds with less than 50 points \cite{de2013unsupervised}) involves both sparsity and unknown pose variation, which are still intractable tasks for the existing approaches. The only known work that is capable of addressing both concerns (sparsity and rotation) is \cite{zhang2017active}. The latter work recognizes sparse points by a simple histogram feature created through bin-counting triangle parameters. However, this method is not scalable to large datasets due to the limited resolution of bins, and also cannot generate point-wise features for segmentation. Instead, we utilize the graph neural network to learn latent representation with high discriminability in an end-to-end fashion, enabling various machine learning applications to be built on top of sparse and rotated point clouds. We summarize our contributions as follows:

\begin{itemize}
    \item Propose a point-wise feature that has invariance towards arbitrary positional and rotational transformations.
    
    \item Propose a graph-based encoder to learn the object level representation that can simultaneously be invariant to positional shift, rotation, and scaling. We show by experiments that the object representation can remain discriminative even for significant sparse points combined with arbitrary rotation and noise jittering.
    
    \item Propose Triangle-Net, an end-to-end deep learning network that utilizes our proposed feature. Our network allows for versatile 3D machine learning tasks to be conducted on point clouds with multifactoral disturbances that cannot be robustly learned by previous methods.
\end{itemize}
}

\section{Related Work}
\subsection{3D Object Recognition}
Existing approaches for point cloud classification mainly include but not limited to: 1) Directly performing classification on point cloud data \cite{qi2017pointnet, qi2017pointnet++, zhao20193d}. 2) Projecting the point cloud data into other formats that can extract features expressively, such as voxelized objects \cite{maturana2015voxnet, brock2016generative}, grid cells \cite{PointGrid, PointwiseConv}, spherical shells \cite{ShellNet}, images taken from multiple view angles \cite{su2015multi, su2018deeper, TDB16a}, or graph representation \cite{PointGCN, DGCNN, LDGCNN, RSCNN}. 3) Learning from hand-crafted features created by point cloud data \cite{ben20183dmfv, zhang2017active, zhang2019rotation, zhang2016triangle}. 4) Building classifiers on top of the learned latent representations using self-supervision e.g. self-reconstruction \cite{wu2016learning, stutz2018learning, dai2017shape}. {Our approach falls into the category of graph based deep learning, because the graph is a particularly suitable technique to utilize the unique structural properties between points.}

\subsection{Position and Orientation Invariance}
Real-world objects can be found in arbitrary shapes and poses and therefore it is necessary to learn the corresponding invariance. However, authors, such as \cite{zhang2019rotation, chen2019clusternet} showed that the variance in orientation may lead to significant performance drops in most mainstream techniques used for point cloud object classification and segmentation. 

{
Various approaches have been developed to alleviate this hurdle. For example, robustness to positional and rotational changes can be either learned \cite{qi2017pointnet, qi2017pointnet++, ben20183dmfv} or manually imported through hand-crafted features \cite{zhang2019rotation, chen2019clusternet, li2020rotation}. However, the learned robustness shows degraded performance when generalized to scenarios where the object rotation is not present in the training set (for example, instances with only rotation around z-axis during training, but in the testing set with arbitrary SO(3) rotations). To add more meaningful variations in training demands more complex network architectures (e.g. more layers, neurons and connections) which increase the training complexity, and thereby limited the ability to generalize to combined variations of many stages. For example, training on a combination of rotation and sparsity can drastically degrade the recognition accuracy both in training and testing, as shown in Sec.~\ref{sec:methods:invarance}. 

A different approach, as used in \cite{zhang2019rotation, chen2019clusternet, li2020rotation} exploits local features with rotational invariance, but all of them are statistically significant only in dense point clouds. However, when the dense point cloud is available, there is not really a need for learning rotation invariance because pre-processing techniques such as alignment by PCA \cite{PCA-Align} can effectively address this problem. In this paper, we focus on addressing the challenge of the point cloud's sparsity with an ambiguous shape envelop. In such cases \cite{zhang2019rotation, chen2019clusternet, li2020rotation, PCA-Align} are not applicable.

\subsection{Robustness to Sparsity}

The two categories of a sparse point cloud representation are: locally sparse and globally sparse. The prior case corresponds to dense point clouds associated with low density regions. This category has been well-studied by either utilizing shape completion \cite{rosales2018gpatlasrrt, yang2017foldingnet, yuan2018pcn} or exploiting local features in regions that are not sparse \cite{ScanObjectNN}. 

The second case of category is a more general case, in which local signatures cannot help with object classification. Therefore, the results can only have been conditioned on the sparse points only. In this paradigm, \cite{qi2017pointnet, qi2017pointnet++, ben20183dmfv, RSCNN, yan2020pointasnl} conducted experiments using a randomly downsampled ModelNet 40 dataset \cite{wu20153d}. However, all evaluation results indicate a significant performance drop when the number of points falls below a threshold. To be specific, methods that exploit local features \cite{qi2017pointnet++, RSCNN, LDGCNN} require statistical significance of correlated points (for example, the points in the $k$-nearest neighbourhood). However, the significance level is reduced as the number of points in a local region decreases. Recently, a new family of approaches based on 2D convolutions on rendered images \cite{su2015multi, su2018deeper} or 3D convolutions \cite{maturana2015voxnet} have been suggested. However, they have been found not suitable when the points are too sparse due to the low correlations between neighboring regions (most regions are void). In addition, none of the above approaches are rotational invariant and therefore the performance would be potentially impacted when the object pose is unknown.
}

\section{Methodology} \label{sec:methods}
{
In this section we explain our method on object recognition, which is robust to point sparsity, positional shifts, scaling and arbitrary rotations. To effectively extract the spatial relationship between points, we utilize a hypergraph based feature proposed in Sec.~\ref{sec:methods:descriptors}-\ref{sec:network}. We then integrate the feature into a deep learning architecture in Sec.~\ref{sec:arch}.
}

\subsection{Graph Representation}\label{sec:methods:descriptors}

For a point cloud denoted by $\mathbf{X}=\left\{\mathbf{x}_{1}, \ldots, \mathbf{x}_{n}\right\}$ and its corresponding surface normals $\mathrm{S}=\left\{\mathbf{s}_{1}, \ldots, \mathbf{s}_{n}\right\}$, an undirected hypergraph $G=(\mathcal{V}, \mathcal{E})$ can be built to represent geometric features; where $\mathcal{V}$ are the vertices that corresponds to the points; and $\mathcal{E}$ are the edge set of the graph, which contains the spatial relationship between points.

We will first explain the process for building the edges representation. Inspired by \cite{HGNN}, we use hypergraph for connecting more than 2 points at once, which gains two advantages: 1) The extracted features have higher dimensions, making each feature to be distinctive from others. 2) A larger number of edges can be constructed when compared to those obtained by only connecting two nodes. This contributes to the statistical significance, which is crucial when the point cloud is sparse.

\begin{figure*}[htb]\centering%
\vspace{-2mm}
\subfigure[]{\includegraphics[width=5cm, angle=0]{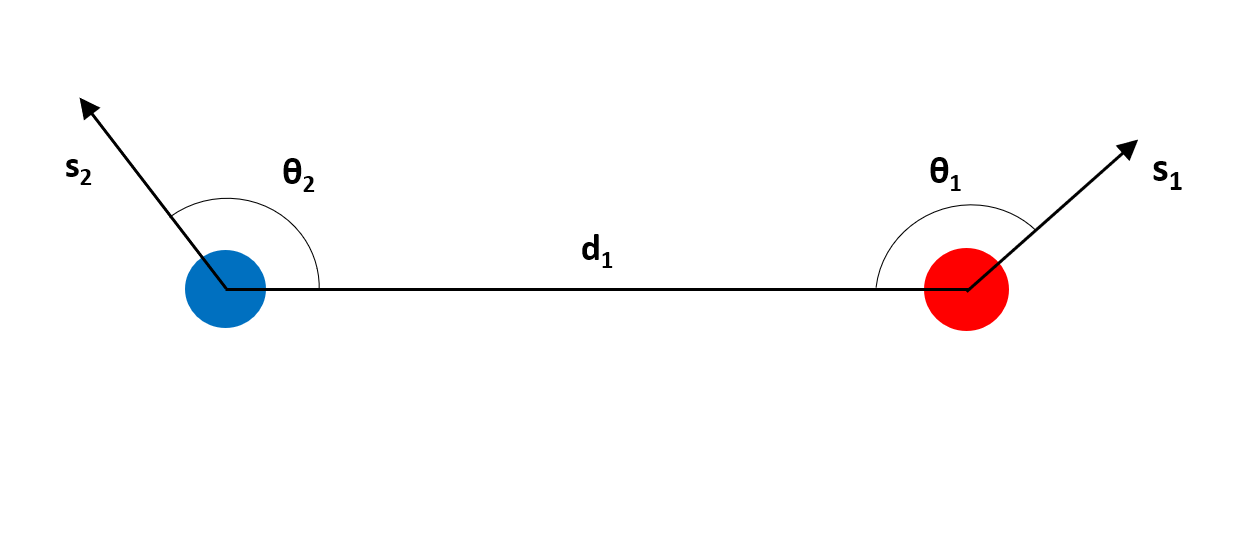}}
\subfigure[]{\includegraphics[width=5cm, angle=0]{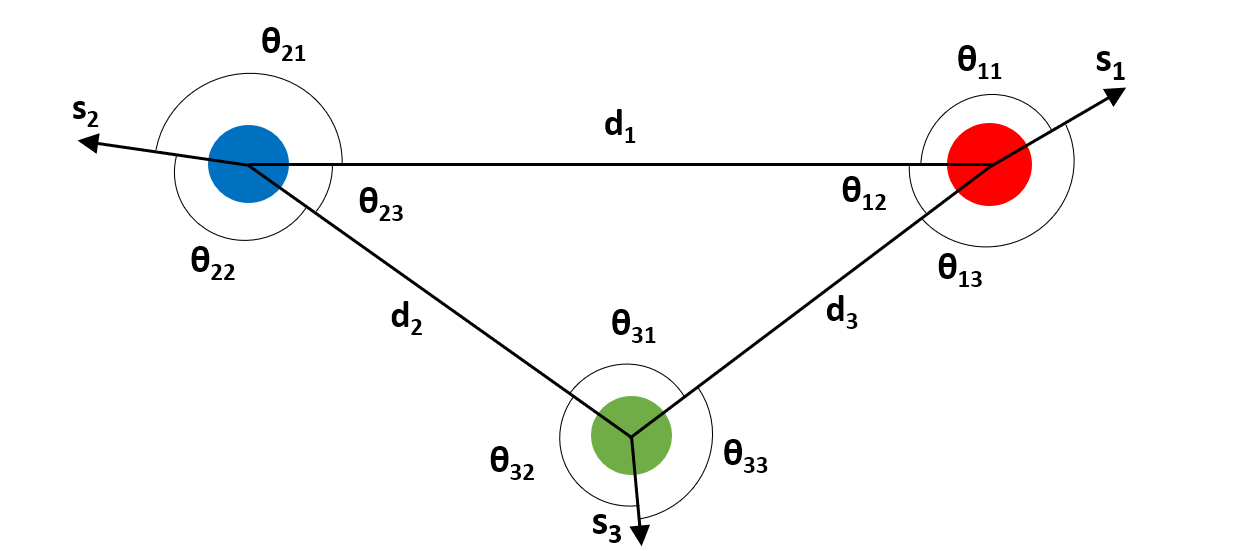}}
\subfigure[]{\includegraphics[width=5cm, angle=0]{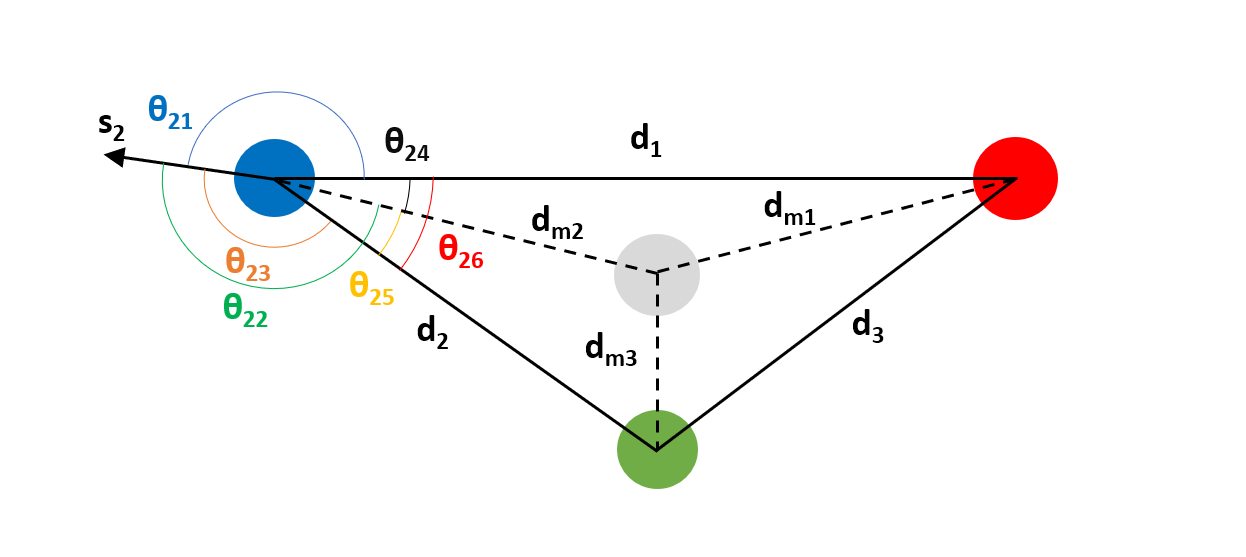}}
\caption{We propose three functions of feature extraction: (a) $\mathcal{D}_A$ , which can be constructed using only 2 points with the attached surface normal vectors.  (b) $\mathcal{D}_B$ that can be constructed using 3 points with surface normal vectors. (c) $\mathcal{D}_C$, which is built on the top of  $\mathcal{D}_B$  but has more pre-computed information.}
\label{descriptors}   
\vspace{-5mm}
\end{figure*}

Fig.~\ref{descriptors} (a) introduces, for the first time, a feature function that utilizes two points $\mathbf{x}_1$, $\mathbf{x}_2$ and its surface normal $\mathbf{s}_1$, $\mathbf{s}_2$. We denote this function as $f(\mathbf{x}_1, \mathbf{x}_2)$, which serves as a building block of our hypergraph representation. The function is given as:
\begin{equation}
\mathcal{D}_A =f(\mathbf{x}_1, \mathbf{x}_2) = \left[d_{\mathbf{x}_1, \mathbf{x}_2},\: \theta_{\mathbf{z}_{12}, \mathbf{s}_1},\: \theta_{\mathbf{z}_{21},\mathbf{s}_2},\: \theta_{\mathbf{s}_1,\mathbf{s}_2}\right]
\end{equation}
Where $\left[ \cdot \right]$ is the concatenate function, $\mathbf{z}_{12} = \mathbf{x}_1 - \mathbf{x}_2$, $d_{12} =\|\mathbf{z}_{12}\|_2$, and $\theta_{\mathbf{u},\mathbf{v}} = arccos{\frac{\left<\mathbf{u}, \mathbf{v}\right>}{\|\mathbf{u}\|_2 \cdot \|\mathbf{v}\|_2}}$. At most $\mathrm{P}_n^2$ non-repetitive $\mathcal{D}_A$ features can be constructed for a point cloud of size $\mathrm{n}$.

Based on the feature function, our proposed type B hyperedge is defined as $\mathcal{D}_B$: 
\begin{equation}
\begin{split}
\mathcal{D}_B\left(\mathbf{x}_1, \mathbf{x}_2, \mathbf{x}_3\right) = & [ f\left(\mathbf{x}_1, \mathbf{x}_2\right),f\left(\mathbf{x}_1, \mathbf{x}_3\right), f\left(\mathbf{x}_2, \mathbf{x}_3\right),\\ & \theta_{\mathbf{z}_{12},\mathbf{z}_{13}},\; \theta_{\mathbf{z}_{21},\mathbf{z}_{23}},\; \theta_{\mathbf{z}_{31}, \mathbf{z}_{32}} ]
\end{split}
\end{equation}

which corresponds to the illustration of Fig.~\ref{descriptors} (b). Note that $\mathcal{D}_B$ not only includes three feature functions but also emphasizes the superimposed spatial relationship between them, meaning the extra elements  $\theta_{\mathbf{z}_{12},\mathbf{z}_{13}}$, $\theta_{\mathbf{z}_{21},\mathbf{z}_{23}}$, $\theta_{\mathbf{z}_{31}, \mathbf{z}_{32}}$. This shows that hyperedges can be used for building highly discriminative feature sets. 

We found that, through experimentation, by including extra hand-crafted features, the accuracy can be improved while reducing convergence time. Therefore, we also propose a feature function $\mathcal{D}_C$, as illustrated at Fig.~ \ref{descriptors} (c). 
A center point between three vertices $\mathbf{x}_m = \frac{1}{3}\left(\mathbf{x}_1+\mathbf{x}_2+\mathbf{x}_3\right)$ is computed, and then additional angular and distance values can be extracted. The math description is given as:
\begin{equation}
\begin{split}
\mathcal{D}_C\left(\mathbf{x}_1, \mathbf{x}_2, \mathbf{x}_3\right) =  
[& \mathcal{D}_B\left(\mathbf{x}_1,\mathbf{x}_2,\mathbf{x}_3\right), \\&
d_{\mathbf{x}_1, \mathbf{x}_m}, \theta_{\mathbf{z}_{1m}, \mathbf{z}_{12}}, \theta_{\mathbf{z}_{1m}, \mathbf{z}_{13}}, \\& 
d_{\mathbf{x}_2, \mathbf{x}_m}, \theta_{\mathbf{z}_{2m}, \mathbf{z}_{21}}, \theta_{\mathbf{z}_{2m}, \mathbf{z}_{23}}, \\&
d_{\mathbf{x}_3, \mathbf{x}_m}, \theta_{\mathbf{z}_{3m}, \mathbf{z}_{31}}, \theta_{\mathbf{z}_{3m}, \mathbf{z}_{32}}
 ]
\end{split}
\end{equation}

{
Robustness to point sparsity can be achieved when the triangle's vertices $\mathbf{x}_1, \mathbf{x}_2, \mathbf{x}_3$ are chosen independently to the point cloud's local density. For example, when $\mathbf{x}_1$ is given, $\mathbf{x}_2$ and $\mathbf{x}_3$ can be chosen by either Farthest Point Sampling or uniform sampling so that our method can exploit the global level geometric relationship.}

An extra robustness, scale invariance can be achieved after introducing the scale normalization. To be specific, for each distance entry in $\mathcal{D}_C\left(\mathbf{x}_1, \mathbf{x}_2, \mathbf{x}_3\right)$, we divide it by $max\left(d_{\mathbf{x_1,x_2}}, d_{\mathbf{x_1,x_3}}, d_{\mathbf{x_2,x_3}}, d_{\mathbf{x_1,x_m}}, d_{\mathbf{x_2,x_m}}, d_{\mathbf{x_3,x_m}}\right)$, resulting the largest distance to be 1. For an object scaled by factor $\alpha$, $\mathcal{D}_C\left(\alpha\mathbf{x}_1, \alpha\mathbf{x}_2, \alpha\mathbf{x}_3\right)$ is strictly equal to $ \mathcal{D}_C\left(\mathbf{x}_1, \mathbf{x}_2, \mathbf{x}_3\right)$ after scale normalization. However, this normalization is optional because object size is also a prior for recognition, while losing scale information is detrimental to the performance.

\subsection{Proof of Invariance}
It can be proved that all $\mathcal{D}_A$, $\mathcal{D}_B$, $\mathcal{D}_C$ proposed above are invariant to arbitrary SO(3) transformations without loss of generality. We denote the rotation transformation as $\mathbf{R}$ and the positional transition vector as $\mathbf{t}$. The SO(3) transformation does not change the distance between points.
\begin{equation}
\begin{split}
d_{\mathbf{x}_1, \mathbf{x}_2} =& \|\mathbf{x}_1 - \mathbf{x}_2\|_2 = \|(\mathbf{R}\mathbf{x}_1 + \mathbf{t}) - (\mathbf{R}\mathbf{x}_2 + \mathbf{t})\|_2 = \\&  \|\mathbf{R}(\mathbf{x}_1 - \mathbf{x}_2)\|_2
\end{split}
\end{equation}

The angle $\theta_{\mathbf{u}, \mathbf{v}}$ between vectors $\mathbf{u =x_2-x_1}$ and $\mathbf{v=x_3-x_1}$ are also invariant to rotation, because:
\begin{equation}
\cos \theta=\frac{\left<\mathbf{u}, \mathbf{v}\right>}{|\mathbf{u}||\mathbf{v}|} = \frac{\mathbf{uR^T} \mathbf{Rv}}{|\mathbf{Ru}||\mathbf{Rv}|} = \frac{\left<\mathbf{u^*}, \mathbf{v^*}\right>}{|\mathbf{u^*}||\mathbf{v^*}|}
\end{equation}
where $\mathbf{u^*} = (\mathbf{Rx}_2 + \mathbf{t}) - (\mathbf{Rx}_1 + \mathbf{t})$ and $\mathbf{u^*} = (\mathbf{Rx}_3 + \mathbf{t}) - (\mathbf{Rx}_1 + \mathbf{t})$. The invariant property can be generalized to surface normal vectors since a surface normal vector $\mathbf{s}_1$ can be rewritten as $(\mathbf{x}_1+\mathbf{s}_1) - \mathbf{x}_1$. 

In brief, since all entries in $\mathcal{D}_A$, $\mathcal{D}_B$, or $\mathcal{D}_C$ are either distance between two points $\mathbf{x}_1$, $\mathbf{x}_2$, or angular value between two vectors, the extracted feature is invariant to SO(3).\hfill $\square$

\subsection{Hyperedge Convolution}\label{sec:network}
Given the hyperedge features being extracted, we leverage on graph aggregation to get the latent feature representation. The aggregate function $\mathcal{A}\left(\mathbf{x}_i\right)$ of point $\mathbf{x}_i$ is given as:
\begin{equation}
\centering
\mathcal{A}\left(\mathbf{x}_i\right)=\max_{j, k \in \mathcal{E}} H_{\Theta}\left(\mathcal{D}_m\left(\mathbf{x}_{i}, \mathbf{x}_{j}, \mathbf{x}_k\right)\right) 
\end{equation}
Where $H_{\Theta}$ is a mapping function parameterized by $\Theta$ implemented by a deep neural network. $\mathcal{D}_{m, m \in \left\{A, B, C\right\}}$ is the proposed feature function described earlier in Sec.~\ref{sec:methods:descriptors}. The dimension-wise $\text{max}$ function is used to aggregate all the transformed features that are correlated with point $\mathbf{x}_i$. $\mathcal{A}\left(\mathbf{x}_i\right)$ is the point-level feature that can then be used for tasks such as point segmentation. Note that only a partial number of features can be extracted when the point cloud size is large due to the computational complexity. Therefore, we only use a feature subset of size $\mathcal{F}$ created by random sampling.

To get the global representation of the whole object, we aggregate all the latent features corresponding to all points in the sampled point cloud. The aggregation can be accomplished by another max function. The global feature from aggregation is then written as:
\begin{equation}
\centering
\mathbf{\mathcal{A}\left(\mathbf{x}_1, ..., \mathbf{x}_n\right)}=\max_{i \in \mathcal{E}} \mathcal{A}\left(\mathbf{x}_i \right)
\end{equation}
Note that the max aggregation function also relaxes the permutation restriction over the input points \cite{qi2017pointnet}. 

\subsection{Network Architecture} \label{sec:arch}

{
We integrate the feature extraction (Sec.~\ref{sec:methods:descriptors}), point/global feature aggregation (Sec.~\ref{sec:network}), and mapping function $H_{\Theta}$ into an end-to-end architecture in Fig.~\ref{paper_main_x}. The mapping function $H_{\Theta}$ is implemented as a neural network, an efficient structure that extracts the feature progressively. We show the number of neurons in each layer in Fig.~\ref{paper_main_x}. Our goal is to utilize deep features that are known to be capable of reducing the inductive bias \cite{lee2015deeply}.  Neurons in each layer receive the concatenated feature from the previous layer as well as the graph feature in Sec.~\ref{sec:methods:descriptors}. This design not only benefits the performance by mitigating the gradient vanishing problem through additional shortcuts but also avoids loss of information with deeper structures. The final representation is a summed feature of the deepest feature and the concatenated feature from shallower layers.

The point level feature and global level feature allows for various machine learning tasks to be conducted using sparse point clouds. While these tasks can be learned separately, we leverage multi-task learning to improve the generalization. The multi-task learning rewards the network to learn the underlying data distribution of the input data and also works as a regularization technique to prevent overfitting \cite{MTLsurvey}. We show the network design of each task as follows.

\textbf{Classification}
The classification network takes in the object global feature $\mathcal{A}(\bm{x_1}, ..., \bm{x_n})$ and predicts the likelihood of target categories. An MLP network with 3 hidden layers is adopted, with 512 and 256 units per hidden layer. Each hidden layer is followed by batch normalization, dropout layer with $p=0.3$ and uses ReLU activation function.

\textbf{Part Segmentation}
We implemented the segmentation network as an MLP network that has 3 layers, with 256, 128, 50 output units, respectively.

\textbf{Voxel Reconstruction}
The voxel reconstruction was conducted by performing upsampling based on $\mathcal{A}(\bm{x_1}, ..., \bm{x_n})$. The upsampling network has four 3D transposed convolutional layers with 1024, 256, 128, 64 kernels, respectively. The network produces voxels with sizes from $4 \times 4 \times 4$ to $32 \times 32 \times 32$ sequentially, with 2$\times$ upsampling rate per stage.

\begin{figure*}\centering%
\vspace{-1mm}
\includegraphics[width=16cm, angle=0]{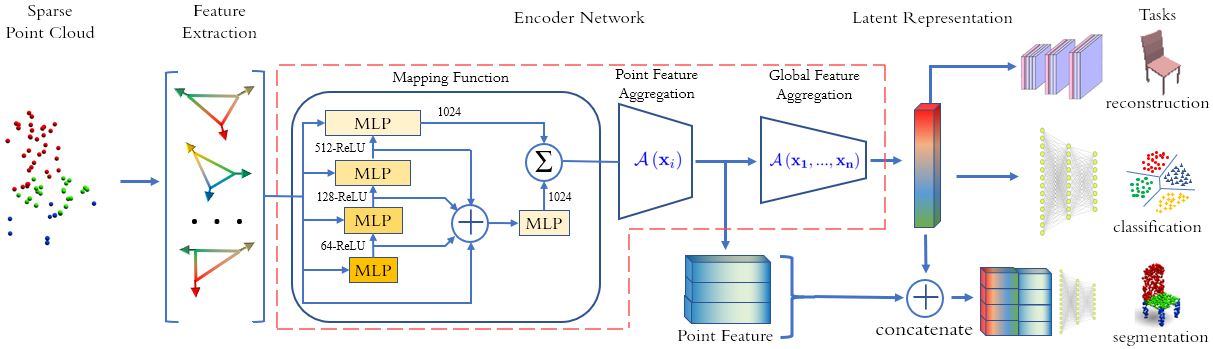}
\caption{Triangle-Net, Our proposed deep learning architecture that extracts features from the parameters of the generated triangles. We leverage on a neural network encoder for extracting point-wise and global features, facilitating classification, segmentation and reconstruction.}
\label{paper_main_x}
\vspace{-5mm}
\end{figure*}
}
\section{Experiments and Results}

\subsection{Experimental Setting} 
{
We evaluate our approach under two scenarios: 1) global sparsity and 2) combined sparsity from both partial and global level. We evaluate two scenarios on ModelNet 40 \cite{wu20153d} and ScanObjectNN \cite{ScanObjectNN} datasets, respectively. ModelNet 40 has 12,311 CAD models in 40 categories while ScanObjectNN has 14,510 scanned point clouds in 15 categories. As oppose to ModelNet 40, objects in ScanObjectNN has a point distribution from real optical scanning that has partial sparsity and even missing regions. For ScanObjectNN, our experiments are conducted on the most difficult variant (PB-T50-RS) without background.

We acquire the data for evaluating global sparsity according to the evaluation protocol in literature \cite{qi2017pointnet,qi2017pointnet++,ben20183dmfv}, by which random downsampling is applied to reduce the number of points without changing the original point distribution pattern. The rotation invariance is evaluated under the same protocol as \cite{zhang2019rotation, chen2019clusternet}, in which a random SO(3) rotation around the object center is applied. 
}

{
\subsection{Ablation Study 1: PointNet's Lack of Robustness
Towards Rotation and Sparsity}
}
Since PointNet is a basic building block for many 3D machine learning algorithms, it is necessary to use an experiment to show that the performance of PointNet will degrade when classifying arbitrarily rotated objects with sparse points. For this, we both trained and tested the PointNet on ModelNet 40 under 3 conditions in different point cloud sizes: a) Point clouds with no rotation b) Only rotate around z-axis or c) Arbitrary SO(3) rotations. 

\begin{table}[b]
\centering
\vspace{-5mm}
\caption{The experiment shows the performance degradation of PointNet (in \%) when using a) No rotation b) Only rotated around z-axis or c) Arbitrary SO(3) rotation (as indicated in different rows) under different point cloud densities (as shown in columns). }
\label{rotation_sparsity}
\setlength{\tabcolsep}{1mm}{
\begin{tabular}{c|cccc}
\hline\hline
No. of points    & 1024  & 256  & 64  & 16 \\\hline
(1) No rotation applied     & 88.51     &   86.89    & 82.49 & 76.40  \\
(2) Rotated around z-axis          & 86.62   &  77.33   & 69.31 & 53.33    \\
(3) Arbitrary SO(3) rotation         & 79.08     &   72.01 & 56.79 & 35.28    \\\hline
Accuracy drop (1)-(3) & 9.43 & 14.88 & 25.80 & 41.12 \\ \hline\hline
\end{tabular}}
\end{table}

The results are shown in Table \ref{rotation_sparsity}. PointNet performs well with dense points cloud under all 3 rotational conditions, as indicated in the first column. The result also shows that PointNet scales well to sparsity when no rotation is applied, as indicated in the first row. However, performance decays as rotation is applied to the data. It can be observed that rotation around the z-axis affects overall performance, and this is further aggravated when arbitrary SO(3) rotations are applied. The T-Net module fails to learn robustness towards SO(3) transformation when the point is extremely sparse.

\subsection{Classification on Sparse and Rotated Points} \label{sec:methods:invarance}
In this experiment, we compare the classification accuracy of our method to other approaches under various point density configurations. The objects in the ModelNet 40 dataset undergo arbitrary SO(3) transformations both in training and testing sets. The results are shown in Table \ref{classification_result}. The last row of the table shows our result with $\mathcal{F}=4096$ features per object. The highest performance is highlighted in \textbf{bold} digits. { Note that 3DmFV and PointNet are known models without a lower boundary for the number of points required. Conversely, PointNet++ and RI-Conv have a lower boundary for the number of points required, thus, is not applicable to cases with fewer than 64 points.  

Last, PointNet was tested under two conditions. The first relies on vanilla PointNet$^1$ which was trained with 1024 points with random input dropout \cite{qi2017pointnet++}. Second, PointNet$^2$ was trained and tested using the same number of points. The results indicate that both PointNet models fail to generalize sufficiently well to sparse point clouds. From Table~\ref{classification_result}, we can see that 3DmFV \cite{ben20183dmfv} can perform better than PointNet when the input points are sparse. However,  the accuracy decays with the point cloud size. We compared our algorithm with RI-CONV \cite{zhang2019rotation} as well, an architecture that has rotational invariance and therefore is robust to SO(3) transformation. However, we found that it is not sufficiently robust to sparsity and displays a performance drop with an increase in the point size.} Other recent methods such as KCNet\cite{KCNET}, KPConv\cite{KPCONV}, require a higher lower boundary for the minimum number of points, and therefore we found not useful to compare with. 

The above comparisons show that our approach can outperform others by a large margin when points are sparse. Conversely, the advantage is not significant when using dense points. We believe this is mainly due to 2 reasons. 1) Part of relative positional information between points is discarded, as each feature is constructed using 3 points rather than all points. 2) When the point cloud is dense, there is an immense number of triangles that can be constructed (e.g. $1 \times 10^9$ possible triangles when using 1024 points) and, in that case, the subset of triangles chosen by our method may be sub-optimal to represent the object of interest.

\begin{table}[H]
\centering
\vspace{-2mm}
\caption{Comparison of ModelNet 40 classification accuracy (in \%) on both globally dense and sparse points under arbitrary SO(3) rotation. Our algorithm shows the advantage when points become sparse.}
\vspace{5pt}
\label{classification_result}
\setlength{\tabcolsep}{1mm}{
\footnotesize
\begin{tabular}{c|cccc|cccc}
\hline\hline
              & \multicolumn{4}{c|}{Dense}       & \multicolumn{4}{c}{Sparse} \\\cline{2-9}
Num of points & 1024  & 512  & 256  & 128        & 64    & 32    & 16   & 8   \\\hline
PointNet$^1$\cite{qi2017pointnet}    & 73.09      & 72.67    & 64.48    & 39.93          & 21.08     & 9.79     & 2.65    & 2.07   \\
PointNet$^2$\cite{qi2017pointnet}    & 79.08     & 75.14    & 72.01    & 72.64          & 56.79     & 48.34     & 35.28    & 23.91   \\
PointNet++\cite{qi2017pointnet++}    & 84.76     & 83.87    & 83.31    & 78.60          & N/A     & N/A     & N/A    & N/A   \\
3DmFV\cite{ben20183dmfv}         & 86.63    & 85.69    & 84.70    & 82.32          & 76.56     & 63.45     & 42.26    & 23.68\\  
RI-CONV\cite{zhang2019rotation}       & 86.5     & 84.4    & 80.8    & 76.0          & N/A     & N/A     & N/A     & N/A    \\
Ours          & \textbf{86.66}     & \textbf{85.73}    & \textbf{85.32}    & \textbf{83.41}          & \textbf{81.53}     & \textbf{79.28}     & \textbf{70.35} & \textbf{48.19}   \\\hline\hline
\end{tabular}}
\vspace{-2mm}
\end{table}

{
We observe a similar trend in the benchmark on ScanObjectNN \cite{ScanObjectNN} dataset that has combined sparsity from global and partial regions, as in Table~\ref{classification_result2}. Our approach shows advantages in all point densities when objects are under arbitrary SO(3) rotations. But when SO(3) rotation and sparsity were not applied, DGCNN \cite{DGCNN} outperformed both PointNet and our approach. We also notice the DGCNN has the most drastic accuracy drop among 3 approaches when the resolution decreases, while ours drops most gracefully among them. In this context, we believe that DGCNN is benefited from the ``EdgeConv'' operation that exploits the local information from the dense parts when the rest is less informative, but this operation may not be effective when points become globally sparse, as evidenced by \cite{retr1, retr2}.}
\begin{table}[H]
\centering
\caption{Comparison of classification accuracy (in \%) on ScanObjectNN dataset. Our algorithm shows the advantage when combined variations are applied. Otherwise DGCNN performs the best.}
\label{classification_result2}
\vspace{5pt}
\vspace{-1mm}
\setlength{\tabcolsep}{1mm}{
\begin{tabular}{c|ccc|ccc}
\hline\hline
              & \multicolumn{3}{c|}{w/o SO(3)}       & \multicolumn{3}{c}{SO(3)} \\\cline{2-7}
Num of points & 2048 & 256 & 32       & 2048 & 256 & 32  \\\hline
PointNet\cite{qi2017pointnet}    & 74.4 & 73.73 & 69.91 & 67.38 & 64.92 & 54.85  \\
DGCNN\cite{DGCNN}    & \textbf{81.5} & \textbf{78.7} & \textbf{70.7} & 71.58 & 69.6 & 55.4   \\
Ours    & 73.77 & 71.82 & 70.16 & \textbf{73.77} & \textbf{71.82} & \textbf{70.16}     \\\hline\hline
\end{tabular}}
\vspace{-1mm}
\end{table}

\subsection{Segmentation on Sparse and Rotated Points}

We conducted a part segmentation experiment based on the ShapeNet part dataset \cite{ShapeNetPart}. This dataset has 14,006 train samples and 2,874 test samples that belong to 16 object categories. Each point was annotated with a label, with 50 types of labeled parts in total. The segmentation task is to predict the label for each point conditioned on both point feature $\mathcal{A}\left(\mathbf{x}_i\right)$ and global feature $\mathbf{\mathcal{A}\left(\mathbf{x}_1, ..., \mathbf{x}_n\right)}$. 

We built the segmentation network on top of our learned representation. For a fair comparison with PointNet, we used a network that has the same classifier head as \cite{qi2017pointnet}. While PointNet can use an arbitrary number of input points, DGCNN requires at least $k$ points for $k$-nearest neighbourhood search. For experiments with 1024, 64, 16 and 8 points, we set $k$ as 20, 16, 8, 4 respectively. The result is shown in Table~\ref{tbl:seg}. Note that our reported IoU value is the IoU averaged over all instances in the test set. 

Our approach also shows an advantage in the segmentation task when the point cloud is sparse. We outperform PointNet and DGCNN with only 8 or 16 points. However, as expected, the IoU decreases as the point cloud becomes denser. Because the number of $\mathcal{D}_C$ features used was too small (only 4096) when compared to the point cloud size, the point-level feature did not represent the target category sufficiently well.

\begin{table}[h]
\centering
\caption{The averaged instance IoU of part segmentation experiment under SO(3) rotation and given point cloud size. The result shows that our approach performs better when points are sparse.}
\vspace{5pt}
\label{tbl:seg}
\begin{tabular}{c|cccc}
\hline\hline
No. of points                   &  1024  & 64    & 16    & 8 \\ \hline
PointNet\cite{qi2017pointnet}   &  \textbf{80.52} & \textbf{80.47} & 75.59 & 70.30 \\
DGCNN\cite{DGCNN}               &  80.43 & 77.94 & 69.27 & 64.38 \\
Ours                            &  72.53 & 80.09 & \textbf{78.74} & \textbf{75.83} \\
\hline\hline
\end{tabular}
\vspace{-3mm}
\end{table}

\subsection{Object Retrieval by Shape Similarity}
\begin{figure}[htb]\centering%
\vspace{-2mm}
\subfigure[]{\includegraphics[width=4.1cm, angle=0]{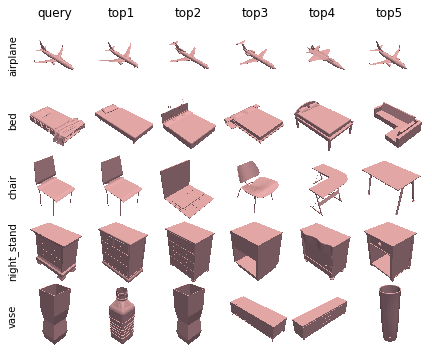}}
\subfigure[]{\includegraphics[width=4.1cm, angle=0]{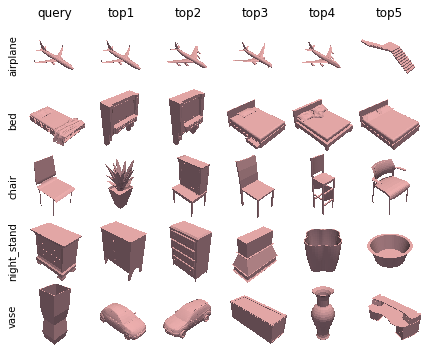}}
\caption{Results of the retrieval operation of our approach (a) vs PointNet model retrieval results (b) using only 16 points under arbitrary SO(3) rotation. }
\label{retrieval_compare}   
\vspace{-4mm}
\end{figure}

Our learned representation can be used as a metric for comparing shape similarity even when the point cloud is sparse and rotated. The experiment below shows the performance of the learned shape similarity metric using only 16 points. Both the query object and the candidate objects are from the test set of ModelNet 40 dataset (i.e. unseen objects). The top 5 similar objects are found within the test dataset using the $k$-nearest neighborhood with the $L^2$ distance metric. The retrieval results of our approach is shown in Fig.~\ref{retrieval_compare} (a). The comparison with PointNet is shown in Fig. \ref{retrieval_compare} (b).

We use retrieval mAP (mean averaged precision) as a metric for a quantitative comparison of our approach to PointNet. Our approach achieves 59.97\% and 56.67\% in top-5 and top-10 retrieval results respectively, outperforming PointNet that achieved 34.80\% and 35.04\% correspondingly. We believe that the boost in performance is from a better discriminative ability of our feature and a better similarity metric learned by object reconstruction. 

\vspace{-1mm}
\subsection{Embedding Analysis}
In order to validate rotational and positional invariance, we observed the feature invariance in the embedding space. First, 3 models were trained with point clouds of size 16 using the feature functions $\mathcal{D}_A$, $\mathcal{D}_B$, $\mathcal{D}_C$ respectively. Then, an arbitrary object $\mathbf{x}_i$ was chosen, and all the non-repetitive features using $\mathcal{D}_A$, $\mathcal{D}_B$, $\mathcal{D}_C$ were created ($\mathrm{P}_{16}^2=240$ features for $\mathcal{D}_A$, $\mathrm{P}_{16}^3=3360$ features for $\mathcal{D}_B$ and $\mathcal{D}_C$) and its embedding vectors were extracted by the encoder. The embedding vector of $\mathbf{x}_i$ remained unchanged after having the input $\mathbf{x}_i$ being subject to random rotation and position transformations, showing that our proposed approach is both rotational and positional invariant.  

\begin{figure*}[ht]\centering%
\subfigure[1024 points]{\includegraphics[width=4cm, angle=0]{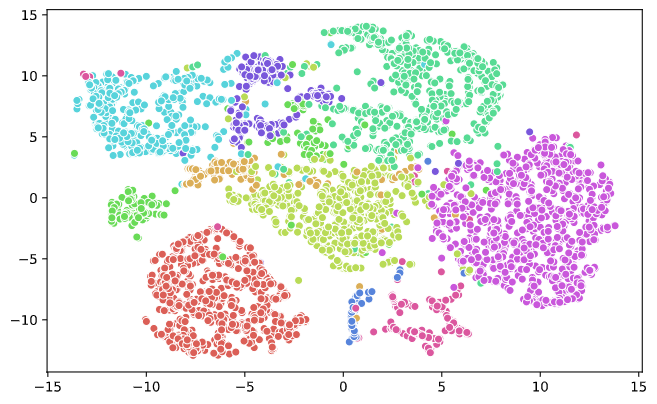}}
\subfigure[64 points]{\includegraphics[width=4cm, angle=0]{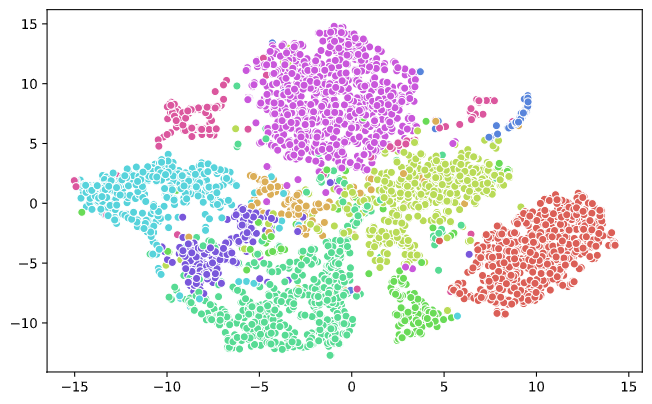}}
\subfigure[16 points]{\includegraphics[width=4cm, angle=0]{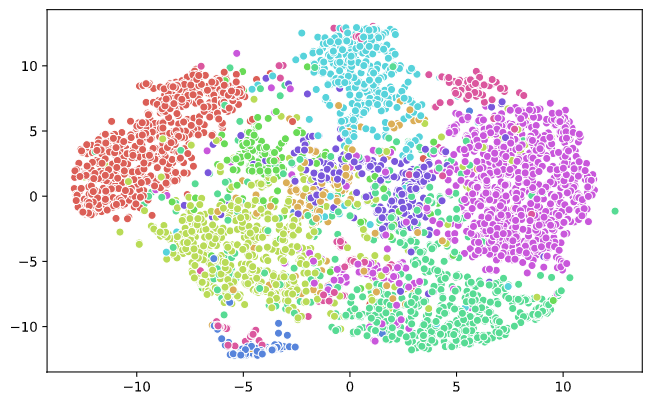}}
\subfigure[8 points]{\includegraphics[width=4cm, angle=0]{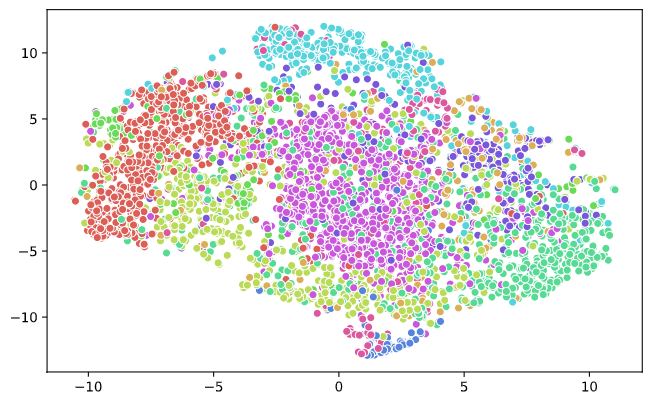}}
\caption{t-SNE visualization of learned embedding space using point cloud of size (a) 1024, (b) 64, (c) 16, (d) 8. The learned features remain to be distinctive even when the point cloud is extremely sparse e.g. 16 or 8 points.}
\vspace{-10pt}
\label{tSNE-exp}                
\end{figure*}

It is observed that the embedding space between categories remained distinctive even when the point cloud was extremely sparse. Further, we compared the data distribution in the feature space under different point cloud size. We used PCA to reduce feature dimensionality from 1024 to 50 and then performed t-SNE in order to create a 2-dimensional visualization. For each plot, 10,000 samples from 10 categories were used. The results shown in Fig. \ref{tSNE-exp}  showcase that our point features are distinctive for all the learned representations. Yet, the margin between categories becomes ambiguous when the points get sparser (e.g. 8 points).  

\vspace{-1mm}
\subsection{Voxel Reconstruction Using Sparse Points}
\begin{figure}[H]%
\centering
\vspace{-2mm}
\includegraphics[width=8cm, angle=0]{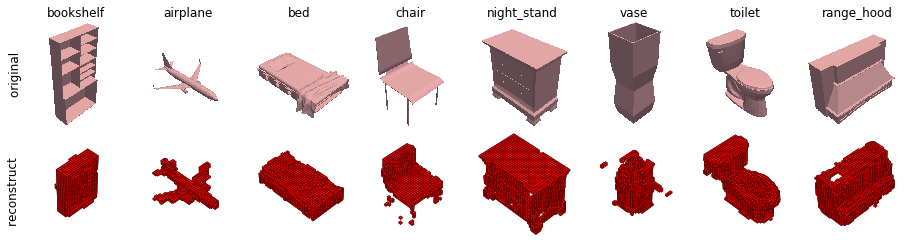}
\caption{Voxel reconstruction using only 16 input points. Even though the input information is very scarce, reasonable reconstruction results can still be achieved.}
\label{reconstruct}    
\vspace{-2mm}
\end{figure}

We show object reconstruction results from the multi-task learning branch using only 16 input points, as Fig.~\ref{reconstruct}.  A voxel is placed when the output (binary Sigmoid function) is larger than 0.2 (instead of 0.5, as the normal Sigmoid case) because we found the network output becomes less "confident" as the input points become sparse. While the reconstruction result resembles the original object, some reconstruction artifacts can still be seen. These include cluttered voxels and inaccurate shape details. We believe this is mainly due to the limited discriminative ability of sparse points.

\subsection{Robustness to Jittering} \label{exp:jittering}

{
Sensor measurements include noisy data due to outliers, errors induced from the tactile sensor (e.g. inaccurate contact normals), or erroneous surface normals estimations (when not accessible from the sensor directly). To answer whether our approach can be robust to such noisy inputs, we conducted noise injection experiments.
}

The noisy surface normal vector $\bm{\hat{s}}$ is generated using the following procedure. First, a random 3D vector $\bm{w_n}$ with a given magnitude $\mathit{m_s}$ is computed: $\bm{v} = m_s\bm{\frac{w_n}{\|w_n\|}}$ where $\bm{w} \sim \mathcal{N}\left(\bm{0}, \bm{I}\right)$. $\bm{v}$ is then being added to the original surface normal $\bm{s}$ and re-normalized to an unit vector: $\bm{\hat{s}=\frac{s+v}{\|s+v\|}}$. 
                
A noisy point $\bm{\hat{p}}$ is generated by adding a Gaussian noise $\bm{w_p}\sim \mathcal{N}\left(\bm{0}, \mathit{m_p}\bm{I}\right)$ to each separate position $\bm{p}$: $\bm{\hat{p}}=\bm{w_p}+\bm{p}$. 

\begin{figure}[H]\centering%
\vspace{-3mm}
\subfigure[position]{\includegraphics[width=4cm, angle=0]{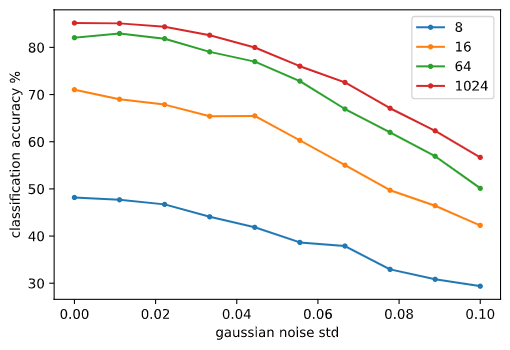}}
\subfigure[surface normal]{\includegraphics[width=4cm, angle=0]{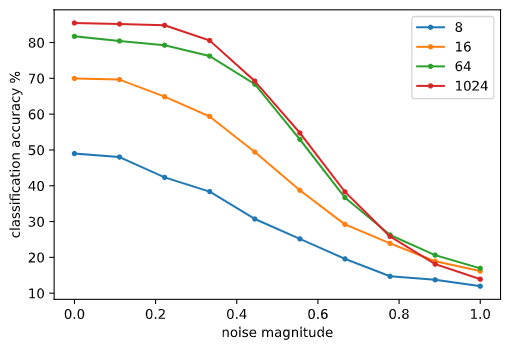}}
\caption{Robustness experiment under jittering in (a) point position (b) direction of surface normal vector, showing that our approach is robust to a wide range of input noise.}
\label{robustness}  
\vspace{-2mm}
\end{figure}

Our approach is validated using the ModelNet 40 classification task with both dense points (1024 points), and sparse points (64, 16, 8 points). For a fair comparison, the model is trained without noise at all. During testing, we inject noise to the surface normal's channel and the position's channel separately. The robustness is evaluated by obtaining the classification accuracy as a function of magnitude $\mathit{m_s}$ and $\mathit{m_p}$, as shown in Fig.~\ref{robustness}. Overall, it can be seen that our network is more robust to noise in both point position's channel and surface normal's channel. For the noisy point position case, our approach only drops 28.5\% when the noise standard deviation reached 10 centimeters in each dimension, while PointNet degrades around 58\% with the same scenario (refer to literature \cite{qi2017pointnet}). For the noisy surface normal case, it shows that robustness varies with point cloud size. When using 8 points (the harshest test scenario), there is only 6.6\% accuracy drop when the random noise component $\bm{v}$ reaches 20\% magnitude of the original surface normal $\bm{u}$. 

\subsection{Scale Invariance}
\begin{figure}[htb]\centering%
\vspace{-2.5mm}
\includegraphics[width=4.5cm, angle=0]{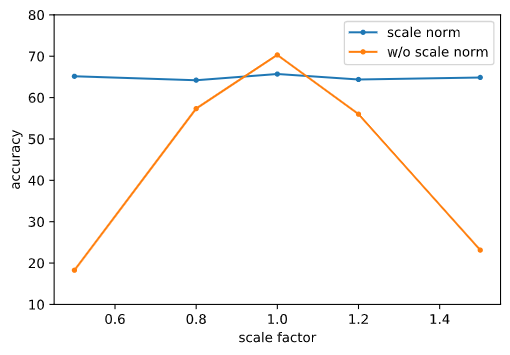}
\caption{Robustness to scaling by the normalization trick (blue curve), otherwise it is sensitive to the object scale (orange curve).}
\label{robustness-scale}      
\vspace{-5.5mm}
\end{figure}

Scale invariance can be achieved with the proposed scale normalization trick. We train 2 models for this experiment. The first model is trained and tested with the scale normalization trick (refer to Sec.~\ref{sec:methods:descriptors}), while the second model is trained and tested without scale normalization. In neither case, data augmentation is used to enhance robustness.
The evaluation is conducted under a combination of 3 variants: arbitrary SO(3) rotation, sparsity (16 points) and scaled by a given ratio from 0.5 to 1.5. The result is shown in Fig.~\ref{robustness-scale}. The blue curve corresponds to the model after the scale normalization trick, showing performance resiliency to changes in scale. Nevertheless, this comes at the price of overall inferior performance than the peak performance achieved by the model without the scale normalization trick (orange curve), as the scale information is lost.

\subsection{Ablation Study 2: Network Components} \label{ablation_study2}
{
\textbf{Inference and Training Time:} Our proposed feature can be computed quickly using parallel computing on a GPU, as shown in Table~\ref{complexity_time}. For ModelNet 40, training with a number of features $\mathcal{F}=4096$ can be completed within 8.5 hours on a single Nvidia Tesla P100 GPU.

\begin{table}[H]
\centering
\vspace{-5pt}
\caption{Model size and inference time under 1024 points.}
\vspace{5pt}
\label{complexity_time}
\small
\begin{tabular}{c|cc}
\hline\hline
Method & \# of parameters & Inference time \\ \hline
PointNet\cite{qi2017pointnet} & 3.5M & 2.1ms \\
DGCNN\cite{DGCNN} & 1.8M & 22.7ms \\
Ours & 2.0M & 6.9ms \\\hline\hline
\end{tabular}
\vspace{-10pt}
\end{table}

\textbf{Number of Features:} The algorithm performance is correlated with the number of generated features $\mathcal{F}$. Table~\ref{numoffeature} shows the ModelNet 40 classification accuracy versus the number of features $\mathcal{F}$ under SO(3) rotation by 1024 points.

\begin{table}[H]
\centering
\vspace{-1mm}
\caption{The ModelNet 40 classification accuracy (in \%) versus number of features $\mathcal{F}$ under 1024 point and SO(3) rotation.}
\vspace{5pt}
\small
\label{numoffeature}
\begin{tabular}{c|cccc}
\hline\hline
Num of $\mathcal{F}$ & 1024 & 2048 & 4096 & 8192\\\midrule
Accuracy          & 85.29     & 86.06 &  86.66 & \textbf{86.99}    \\\hline\hline
\end{tabular}
\vspace{-10pt}
\end{table}

}

We compare several variations of our approach quantitatively using 16 points on ModelNet 40, as shown in Fig.~\ref{ablation_study_variations}. Classification results increase as more information is added to the feature. Accuracies of 60.08\%, 69.04\%, and 69.48\% are achieved using $\mathcal{D}_A$, $\mathcal{D}_B$, $\mathcal{D}_C$ feature functions when only trained on the classification task, and the accuracy is further boosted to 70.35\% when classification is trained together with object reconstruction. The latter scenario corresponds to the highest accuracy we achieved. 

\begin{table}[H]
\centering
\vspace{-2pt}
\caption{ModelNet 40 classification accuracy (in \%) of several variations of our proposed algorithm.}
\label{ablation_study_variations}
\small
\begin{tabular}{c|c|c}
\hline\hline
Descriptor                           & Reconstruction              & Accuracy\\\hline
$\mathcal{D}_A$ feature function                    & No                          & 60.08 \\
$\mathcal{D}_B$ feature function                    & No                          & 69.04\\
$\mathcal{D}_C$ feature function                    & No                          & 69.48\\
$\mathcal{D}_C$ feature function                    & Yes                         & \textbf{70.35}\\
\hline\hline
\end{tabular}
\vspace{-8pt}
\end{table}

\section{Conclusions}
While a rich variety of 3D object recognition methods have been proposed over recent years, very few of them can work on point clouds with a combination of disturbances such as low resolution, unaligned pose, and varied object scale. To address this problem, we evaluated state-of-the-art approaches under arbitrarily rotated sparse point clouds, and found most approaches only achieve limited performance or cannot work under this setting altogether.

{
In this paper, we propose a robust feature extraction method for point cloud that can generate invariant features towards positional, rotational and scaling disturbances. Such type of feature can remain discriminative when the point cloud is of significant sparsity and even being perturbed with noise.
Furthermore, the feature extraction mechanism is integrated into Triangle-Net, a deep neural network that can learn in an end-to-end fashion. Experiments were conducted to show that our learned representation can remain robust to multifactorial variations, and is resilient to jittering, facilitating universal 3D machine learning tasks to be conducted on imperfect measurements and limited resources. 
}

\vspace{-2mm}
\section{Acknowledgement}
This material is based upon work supported by the National Science Foundation under Grant NSF NRI \#1925194. Any opinions, findings, and conclusions or recommendations expressed in this material are those of the author(s) and do not necessarily reflect the views of the National Science Foundation.

The code for this work is released at: \url{github.com/MegaYEye/Triangle-Net.git}

{
\clearpage
\newpage
\small
\bibliographystyle{ieee}
\bibliography{egbib}
}

\end{document}